\documentclass[letterpaper, 10 pt, conference]{ieeeconf}  

\IEEEoverridecommandlockouts                              

\usepackage{graphics} 
\usepackage{epsfig} 
\usepackage{mathptmx} 
\usepackage{times} 
\usepackage{amsmath} 
\usepackage{amssymb}  
\usepackage[T1]{fontenc}
\usepackage{multirow}
\usepackage{subfig}
\usepackage{color}
\usepackage{xcolor}
\usepackage{comment}
\usepackage{tabularx}
\usepackage{multirow}

\newcommand{\figref}[1]{Fig. \ref{#1}}
\newcommand{\tabref}[1]{Table \ref{#1}}

\makeatletter
\def\hlinewd#1{%
\noalign{\ifnum0=`}\fi\hrule \@height #1 \futurelet
\reserved@a\@xhline}

\title{\LARGE \bf
Semi-Supervised Learning with Mutual Distillation for Monocular Depth Estimation}

\author{Jongbeom Baek$^{*}$, Gyeongnyeon Kim$^{*}$, and Seungryong Kim
\thanks{$*$ Joint first authorship}%
\thanks{CVLAB, Korea University, Seoul, Korea}
}

\begin{document}

\maketitle
\thispagestyle{empty}
\pagestyle{empty}

\begin{abstract}
We propose a semi-supervised learning framework for monocular depth estimation. Compared to existing semi-supervised learning methods, which inherit limitations of both sparse supervised and unsupervised loss functions, we achieve the complementary advantages of both loss functions, by building two separate network branches for each loss and distilling each other through the mutual distillation loss function. We also present to apply different data augmentation to each branch, which improves the robustness. We conduct experiments to demonstrate the effectiveness of our framework over the latest methods and provide extensive ablation studies.
\end{abstract}

\section{Introduction}
Monocular depth estimation, estimating a depth map from a single image, can facilitate numerous Computer Vision and Robotics applications, such as scene understanding, SLAM, and autonomous driving~\cite{mur2015orb, tateno2017cnn, zhong2018detect}. 

Early approaches~\cite{eigen2014depth,li2015depth, wang2015designing,kim2016,ummenhofer2017demon} that solve the task with deep Convolutional Neural Networks (CNNs) were formulated in a \textit{supervised} manner, which relies on large ground-truth depth data. But, constructing such data is very costly and labour-intensive~\cite{garg2016unsupervised,xie2016deep3d}. In addition, they have a limited generalization ability to the regions where the ground-truth depths do not exist. To alleviate the reliance on large ground-truth data, \textit{unsupervised}, also called \textit{self-supervised}, learning based methods~\cite{garg2016unsupervised, xie2016deep3d, godard2017unsupervised,luo2018single, godard2019digging} have been proposed, which cast the task as an image synthesis from stereo image pairs or monocular video sequences. Although it turns out that the unsupervised loss is an appealing alternative, but it often leads to blurry results around depth boundaries~\cite{godard2017unsupervised,alvarez2019self}.

To learn monocular depth estimation networks in a \textit{semi-supervised} manner, some methods~\cite{kuznietsov2017semi,amiri2019semi} directly combine the sparse supervised and unsupervised loss functions, but such straightforward approach inherits limitations of both loss functions and accumulates the errors by each loss, which the networks trained with each loss solely may handle.

Meanwhile, some methods~\cite{guo2018learning,tonioni2019unsupervised,cho2019large} attempted to train monocular depth estimation networks through pseudo depth labels from stereo image pairs generated by traditional or pre-trained stereo matching modules~\cite{tonioni2019unsupervised,cho2019large}. To mitigate performance degeneration by inaccurate pseudo depth labels, the confidence of the pseudo depth is also predicted. However, demanding additional stereo matching and confidence estimation modules hinders their applicability.
\begin{figure}[t]
    \begin{center}
    \includegraphics[width=1\linewidth]{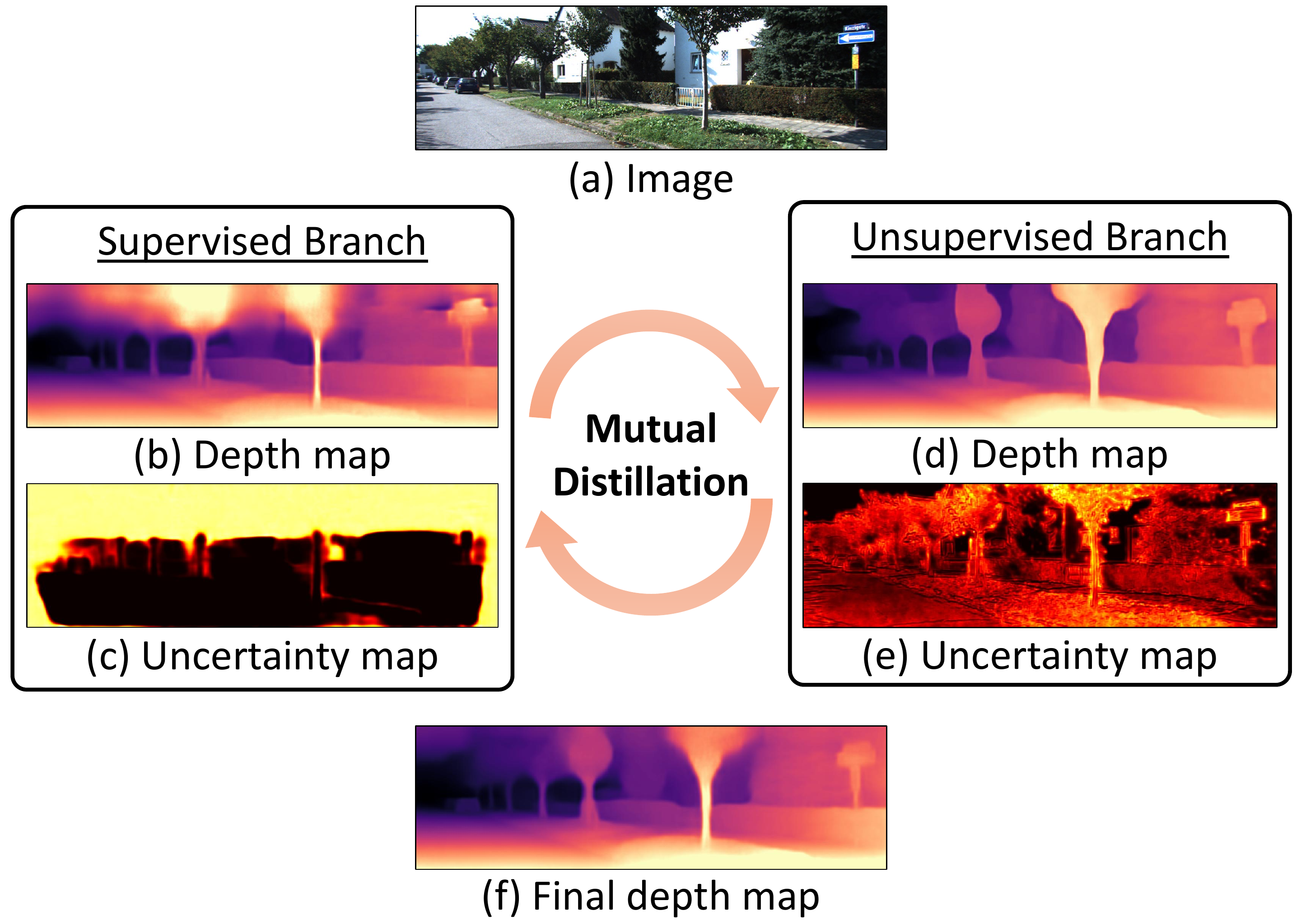}
    \end{center}\vspace{-5pt}
    \caption{\textbf{Illustrations of our approach} that predicts depth and uncertainty for supervised and unsupervised loss functions through two separate networks independently and distil the depth knowledge each other as evolving training. Note that following~\cite{poggi2020uncertainty}, we choose colormap \textbf{magma} for depth and \textbf{hot} for uncertainty (Best viewed with colors).}
    \label{fig_1}\vspace{-10pt}
\end{figure}
\begin{figure*}[t]%
\begin{center}%
\subfloat[][Conventional semi-supervised methods]{\includegraphics[width=0.33\linewidth]{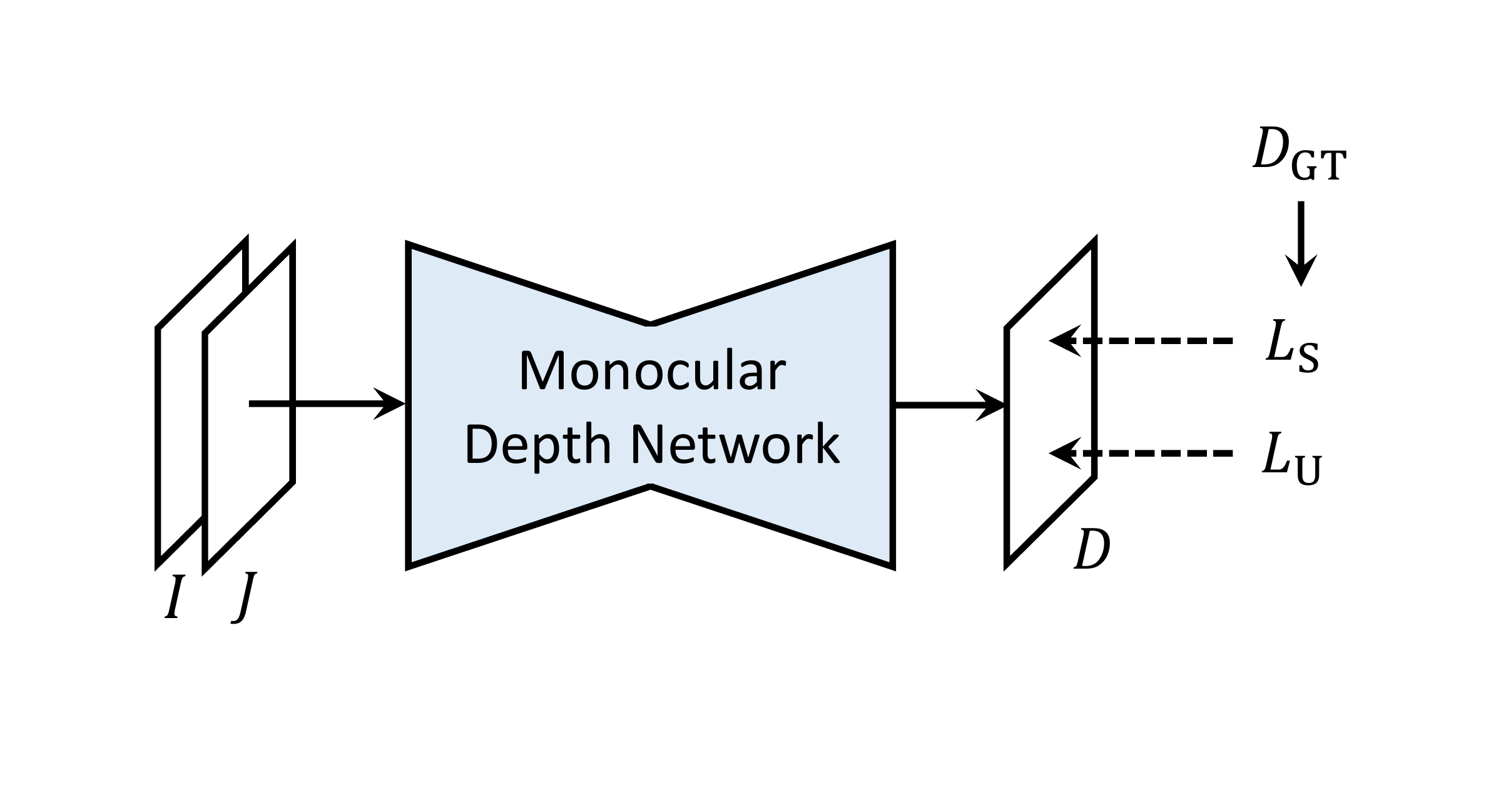}}%
\hfill
\subfloat[][Stereo knowledge distillation methods]{\includegraphics[width=0.33\linewidth]{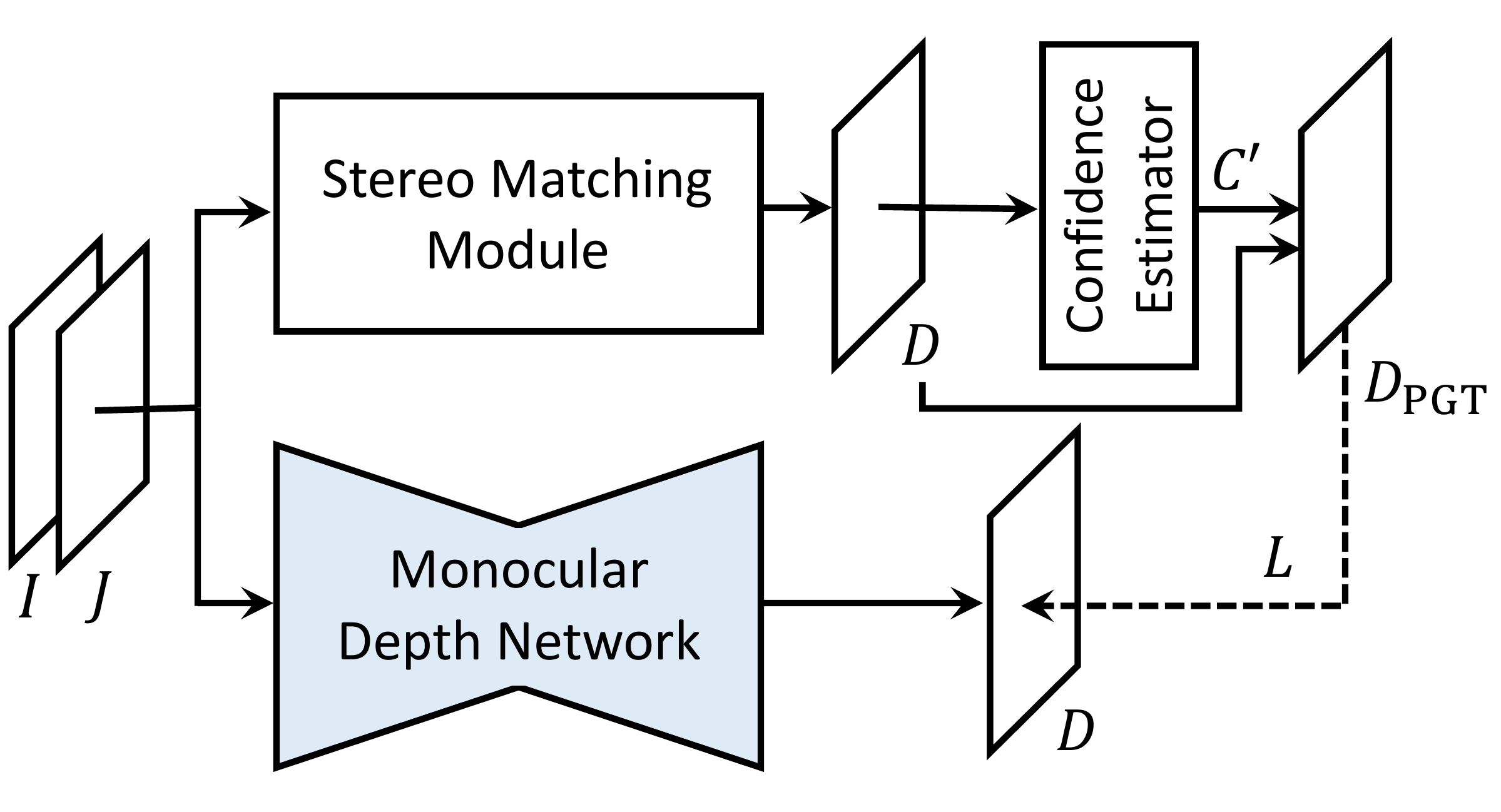}}%
\hfill
\subfloat[][Ours]{\includegraphics[width=0.33\linewidth]{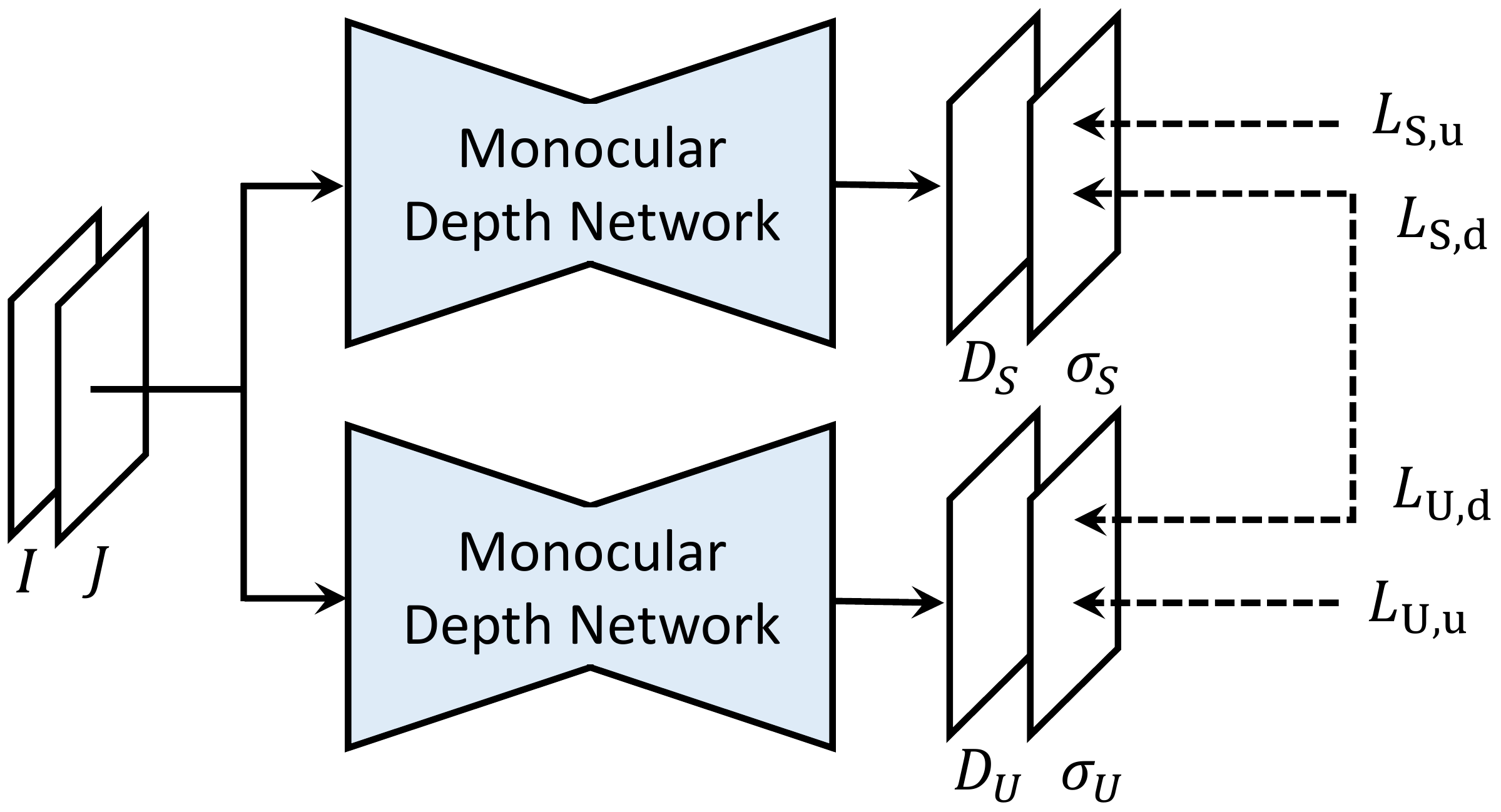}}%
\end{center}\vspace{-10pt}
\caption{\textbf{Motivation:} (a) existing semi-supervised methods~\cite{kuznietsov2017semi,amiri2019semi} that leverage sparse supervised and unsupervised loss functions through a simple summation, which inherit both limitations of the two losses, (b) stereo knowledge distillation frameworks~\cite{pilzer2019refine,tosi2019learning,cho2019large} that distil the depth knowledge by a stereo matching module to the monocular depth estimation networks, which requires additional stereo matching and confidence estimation modules, and (c) our framework that builds separate monocular depth estimation networks for sparse supervised and unsupervised loss functions and distils each other.}%
\label{fig_2}\vspace{-10pt}
\end{figure*}

To overcome the aforementioned limitations, we propose a novel \textit{semi-supervised} learning framework for monocular depth estimation, as in~\figref{fig_1}. To achieve the complementary advantages of both sparse supervised and unsupervised loss functions, we present to build separate networks tailored for each loss, called supervised and unsupervised branches. We train the networks in a probabilistic inference framework to predict both the depth and its uncertainty, so as to distil confident depth knowledge between the branches. To consider the sparsity nature of supervised loss, we introduce unprojected point filtering loss at supervised branch to localize the regions, where ground-truth depths do not exist. By leveraging the proposed mutual distillation loss, defined with depth and confidence maps from each branch, two networks converge to better depth solutions in a mutual and boosting manner. In addition, our framework enables applying different data augmentation to each branch, which improves the robustness. Experiments on standard benchmarks for monocular depth estimation such as KITTI~\cite{geiger2013vision} and Cityscapes~\cite{cordts2016cityscapes} prove the effectiveness of our approach over the latest methods. We also provide an ablation study to validate and analyze components in our approach.

\section{Related Work}
\textbf{Monocular Depth Estimation.}
Eigen et al.~\cite{eigen2014depth} pioneered the approach with deep CNNs for monocular depth estimation. Following~\cite{eigen2014depth}, several methods have been proposed to improve the performance~\cite{li2015depth,wang2015designing,kim2016,laina2016deeper,ummenhofer2017demon}. Traditional methods mentioned above were often formulated in a \textit{supervised} manner that require massive ground-truth depth maps.

To address this limitation, \textit{self-supervised} approaches based on other forms of supervision, i.e., from stereo image pairs or video, have been introduced~\cite{xie2016deep3d,garg2016unsupervised,godard2017unsupervised,zhou2017unsupervised,godard2019digging}. \cite{garg2016unsupervised} used the stereo reconstruction. \cite{godard2017unsupervised} used additional loss to enforce left-right consistency of the predicted disparities. \cite{zhou2017unsupervised} and \cite{godard2019digging} simultaneously learn depth and pose networks from video. Although they seem to be an appealing alternative, the reconstruction loss has problems, e.g., missing objects and over-smoothing boundaries.

To take advantages of both sparse supervised and unsupervised learning methods, \textit{semi-supervised} learning methods have also been presented~\cite{kuznietsov2017semi,amiri2019semi}. \cite{kuznietsov2017semi} directly combined both loss functions. \cite{amiri2019semi} improved performance by applying left-right consistency loss. They inherit limitations of both loss functions, and overcoming this is the topic of this paper.

Meanwhile, as a proxy loss signal, using stereo knowledge for monocular depth estimation is also proposed~\cite{guo2018learning,tonioni2019unsupervised,cho2019large}. Tonioni et al.~\cite{tonioni2019unsupervised} showed the possibility to substitute the ground-truth with proxy labels obtained through traditional stereo matching and confidence measure~\cite{poggi2017quantitative}. \cite{guo2018learning, cho2019large} utilize pseudo labels from pre-trained stereo matching network to train student by distilling the knowledge.  However, requiring additional stereo matching and confidence estimation modules limits their applicability. Unlike these methods~\cite{guo2018learning,tonioni2019unsupervised,cho2019large}, our approach only exploits the monocular depth estimation networks.

\textbf{Self-training.}
Self-training is one of popular semi-supervised learning approaches that encourages a model to follow the pseudo label from the model's prediction itself for entropy minimization~\cite{grandvalet2005semi}, which has been successfully utilized in many tasks~\cite{yalniz2019billion,xie2020self,zoph2020rethinking}. 
However, when the pseudo labels are inaccurate, vanilla methods often produce confirmation bias~\cite{arazo2020pseudo}. To overcome this, \cite{pham2021meta} learns the teacher as well to enhance the student's performance on labeled dataset. Our approach is the first attempt to exploit \textit{self-training} paradigm for monocular depth estimation.

\textbf{Uncertainty Estimation.}
Estimating the uncertainty of predictions is critical for safety-critical applications~\cite{mackay1992practical, welling2011bayesian, gal2016uncertainty}. There are two main types of uncertainty, i.e., epistemic uncertainty and aleatoric uncertainty~\cite{der2009aleatory}. Epistemic uncertainty is in the model, which captures ignorance about the models due to the lack of training data.  Aleatoric uncertainty, on the other hand, captures noise inherent in the environment such as measurement noise. Kendall et al.~\cite{kendall2017uncertainties} studied the benefits of modeling uncertainty in Bayesian deep learning models for vision tasks. For monocular depth prediction, Poggi et al.~\cite{poggi2020uncertainty} studied how to measure the uncertainty for monocular depth estimator. In this paper, we propose to use uncertainty for mutual distillation.

\section{Preliminaries}
Let us denote an image and depth as $I$ and $D$, respectively. Our objective is to learn monocular depth estimation network that takes the image $I$ as input and produces its corresponding depth $D$. To learn the network in a \textit{supervised} manner, the ground-truth depth $D_\mathrm{GT}(i)$, defined at all points $i$, is required, but establishing a large-scale \textit{dense} depth data is very costly and labour-intensive~\cite{liang2018learning}. Instead, \textit{sparse} ground-truth depth labels can be relatively easily acquired, e.g., by the LiDAR sensor, since the LiDAR laser measurements are projected to a sparse subset in an image with a limited amount of scan lines~\cite{kuznietsov2017semi}. Recent research trends thus focus on leveraging the unlabeled data, i.e., adjacent images or the points having no ground-truth depth labels, while mitigating the reliance on labeled data, i.e., sparse depth maps. 

When sparse supervision is available to train the network, we can minimize sparse supervised loss function $L_\mathrm{S}$ between predicted $D$ and sparse ground-truth $D_\mathrm{GT}$ such that
\begin{equation}
    L_\mathrm{S} = 
    \frac{1}{N_D}\sum_{i\in\Omega_D} \|D(i)-D_\mathrm{GT}(i)\|_1,
\end{equation}
where $\Omega_D$ indicates a set of pixels where ground-truth depths are available, and $N_D$ is the number of pixels in $\Omega_D$. Due to sparsity of the ground-truth (e.g., 3\% density in KITTI depth maps~\cite{geiger2012we}), minimizing the loss function $L_\mathrm{S}$ solely cannot guarantee high-precision depth estimation, e.g., especially at sky regions or transparent object regions where the LiDAR sensor does not capture~\cite{uhrig2017sparsity}.
\begin{figure}[t]
    \begin{center}
    \includegraphics[width=1\linewidth]{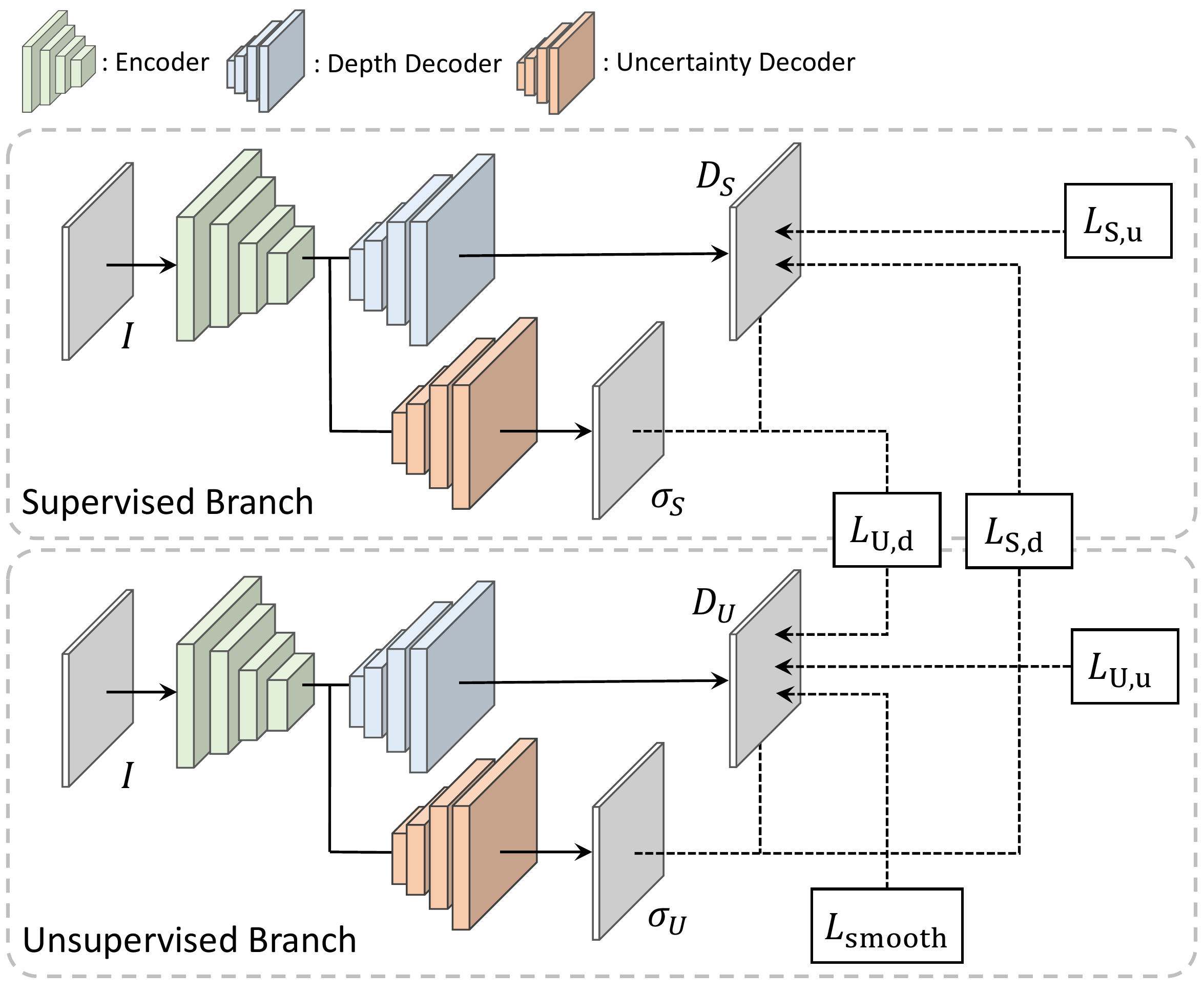}
    
    \end{center}\vspace{-5pt}
    \caption{\textbf{Network configuration.} Our network consists of the same encoder-decoder structure for supervised and unsupervised branches. To identify inaccurate and uncertain depths, we simultaneously predict depth and uncertainty. The separate networks are trained through each loss, and the distillation loss mutually boosts each other.}
    \label{fig_3}\vspace{-10pt}
\end{figure}

Instead of using ground-truth depths for training, \textit{unsupervised} or \textit{self-supervised} learning-based methods~\cite{godard2017unsupervised,zhou2017unsupervised,godard2019digging} formulate the loss function to minimize a photometric reprojection error between adjacent images. Given another image $J$, the relative pose $T$ for $J$ with respect to $I$'s pose is used to warp $J$ towards $I$ such that
\begin{equation}
    I' = J<\mathrm{proj}(D,T,K)>,
\end{equation}
where $\mathrm{proj}$ are the resulting 2D coordinates of the projected depths $D$ of $I$ in $J$, $K$ is an intrinsic matrix, and $<\cdot>$ is the sampling operator~\cite{jaderberg2015spatial}. Then, the unsupervised loss function $L_\mathrm{U}$ is defined between $I$ and $I'$ such that
\begin{equation}
    L_\mathrm{U} = \frac{1}{N} \sum_{i\in\Omega} \mathrm{pe}(I(i),I'(i)),
\end{equation}
where $\Omega$ indicates all the pixels, $N$ is the number of pixels in $\Omega$, and $\mathrm{pe}$ is a photometric reconstruction error as
\begin{equation}
    \mathrm{pe}(I,I') = \alpha (1-\mathrm{SSIM}(I,I'))/2 + (1-\alpha) \| I-I' \|_{1},
\end{equation}
where L1 distance and structural similarity (SSIM)~\cite{wang2004image} are used, following~\cite{zhao2015loss}. It seems to be an appealing alternative to the lack of large-scale ground-truth labels, but it often leads to blurry results around depth boundaries and does not consider occluded pixels~\cite{godard2019digging}.
\begin{figure}[t]
\centering
\subfloat[Image]
{\includegraphics[width=0.33\linewidth]{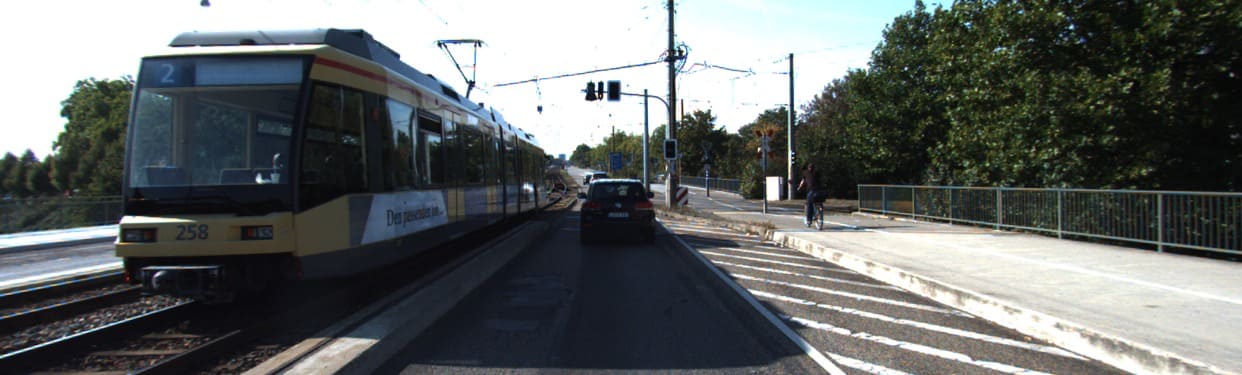}}\hfill
\subfloat[Ground-truth depth]
{\includegraphics[width=0.33\linewidth]{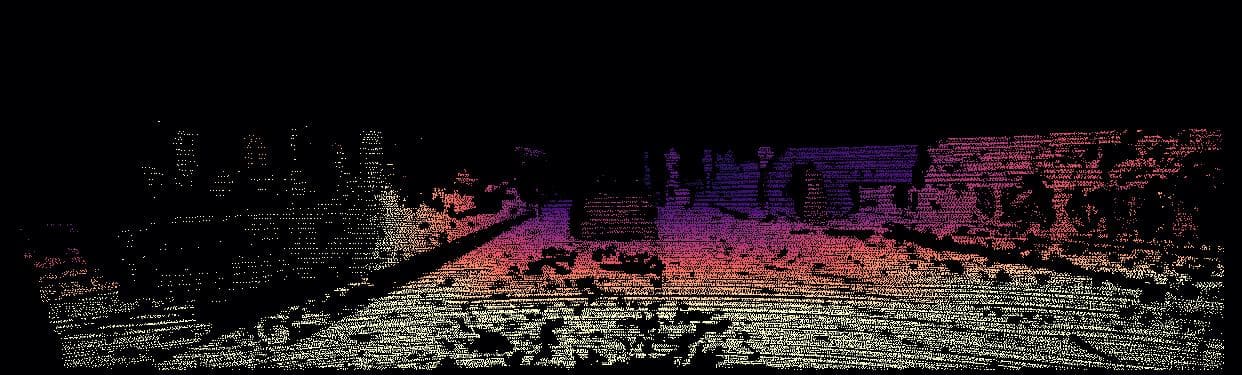}}\hfill
\subfloat[Depth w/ $L'_\mathrm{S,u}$]
{\includegraphics[width=0.33\linewidth]{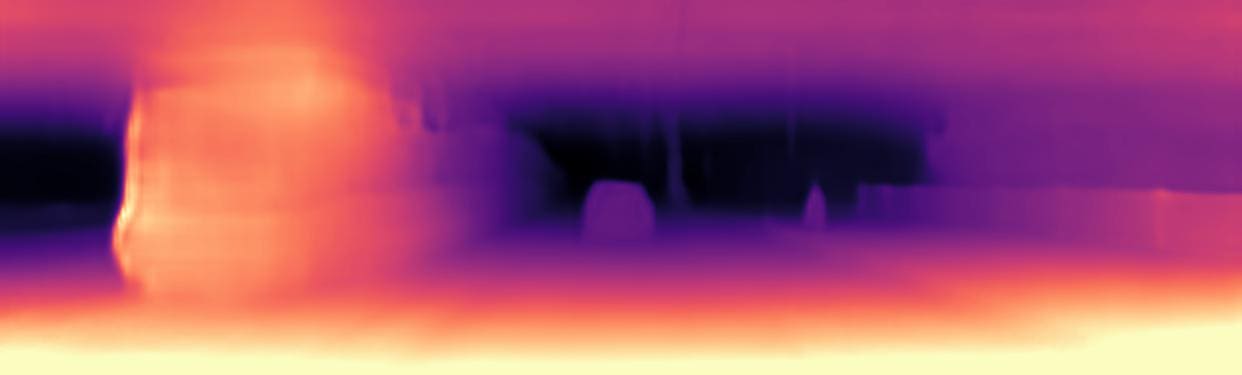}}\hfill
\\ \vspace{-5pt}
\subfloat[Uncertainty w/ $L'_\mathrm{S,u}$]
{\includegraphics[width=0.33\linewidth]{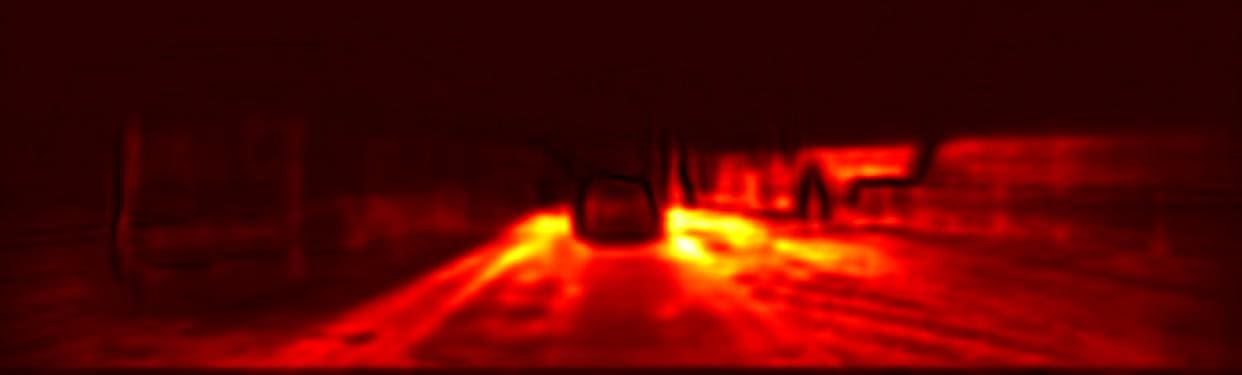}}\hfill
\subfloat[Depth w/ $L_\mathrm{S,u}$]
{\includegraphics[width=0.33\linewidth]{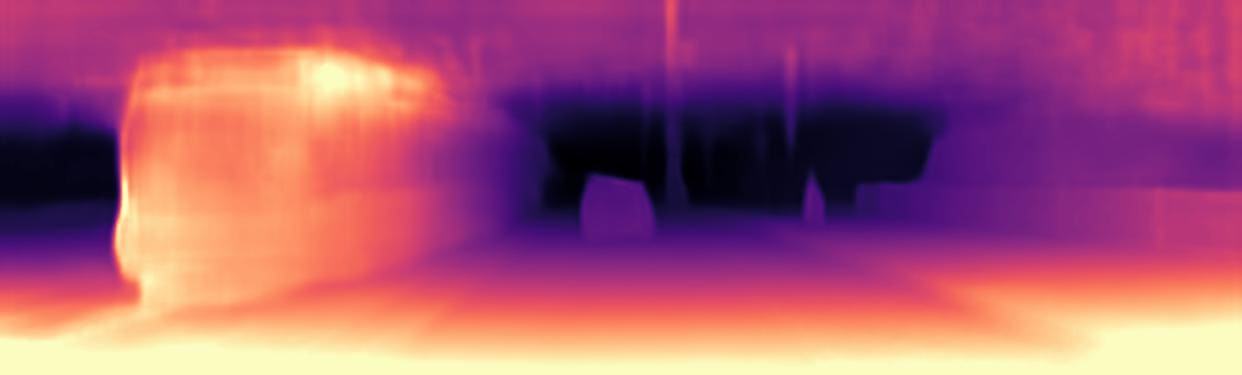}}\hfill
\subfloat[Uncertainty w/ $L_\mathrm{S,u}$]
{\includegraphics[width=0.33\linewidth]{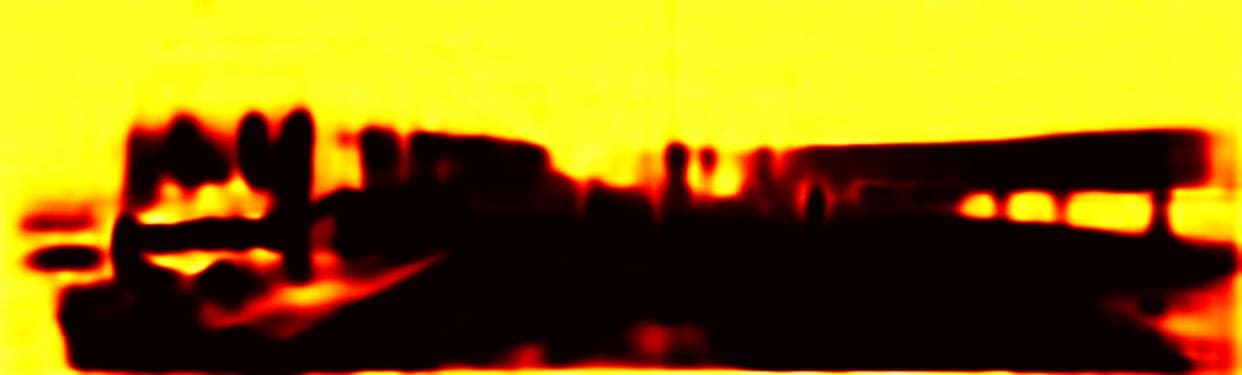}}\hfill
\\ \vspace{-5pt}
\caption{\textbf{Comparison of predicted depth and uncertainty maps using the unprojected point filtering loss:} (a) color image, (b) ground-truth depth, depths and uncertainties by (c), (d) without and (e), (f) with the proposed unprojected point filtering loss, where the latter better captures the region where ground-truth depths do not exist.}
\label{unprojected}
\vspace{-10pt}
\end{figure}

In addition, as suggested in~\cite{godard2017unsupervised}, an depth smoothness term is often incorporated since the depth is not continuous around object boundaries such that
\begin{equation}
    L_\mathrm{smooth} = 
    \frac{1}{N}
    \sum_{i\in\Omega}
    \left(|\partial_{x} D(i)|\mathrm{e}^{-|\partial_{x}I(i)|}
    + |\partial_{y}D(i)|\mathrm{e}^{-|\partial_{y}I(i)|}\right).
\end{equation}

\section{Method}
\subsection{Motivation and Overview}
To learn monocular depth estimation networks in a \textit{semi-supervised} manner, methods~\cite{kuznietsov2017semi,amiri2019semi} directly \textit{combine} the supervised and unsupervised loss functions, $L_\mathrm{S}$ and $L_\mathrm{U}$, with the smoothness loss $L_\mathrm{smooth}$, as illustrated in~\figref{fig_2}(a):
\begin{equation}
    L = L_\mathrm{S} + \lambda L_\mathrm{U} + \lambda_\mathrm{smooth} L_\mathrm{smooth},
\end{equation}
where $\lambda$ and $\lambda_\mathrm{smooth}$ denote weighting parameters. Though simple and straightforward, these methods inherit limitations of both sparse supervised loss function $L_\mathrm{S}$ and unsupervised loss function $L_\mathrm{U}$ in that errors derived from each loss cannot be handled. For instance, it is well known that unsupervised loss function $L_\mathrm{U}$ often leads to blurry results around depth boundaries~\cite{godard2019digging}, and the single network trained both with $L_\mathrm{S}$ and $L_\mathrm{U}$ accumulates such errors, even through a network trained solely with $L_\mathrm{S}$ may recover such depth boundaries well. In addition, as evolving training iterations, the networks become to generate better depth maps, which may be used to train the networks, as done in pseudo-labeling methods~\cite{xie2020self,pham2021meta}, but such loss combination cannot 
be formulated with such pseudo-labeling framework.
\begin{figure*}[t]
\centering
\includegraphics[width=0.165\linewidth]{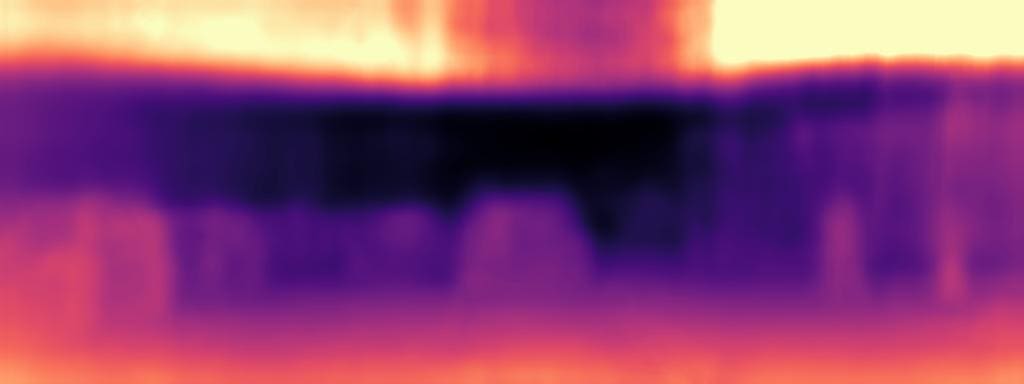}\hfill
\includegraphics[width=0.165\linewidth]{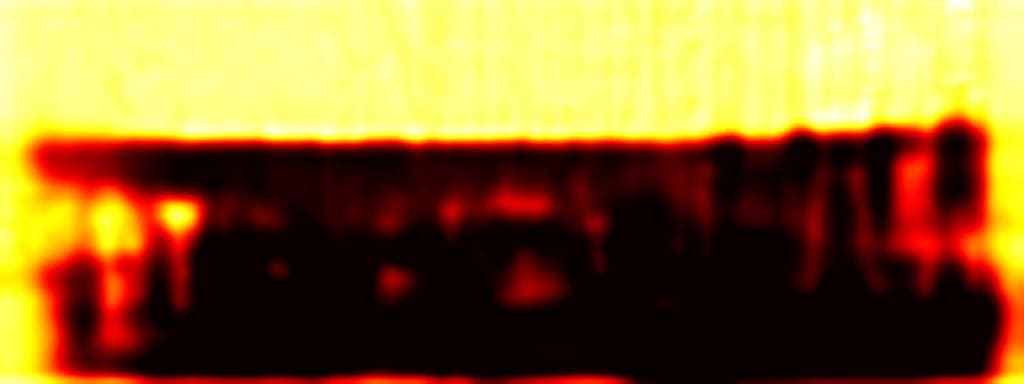}\hfill
\includegraphics[width=0.165\linewidth]{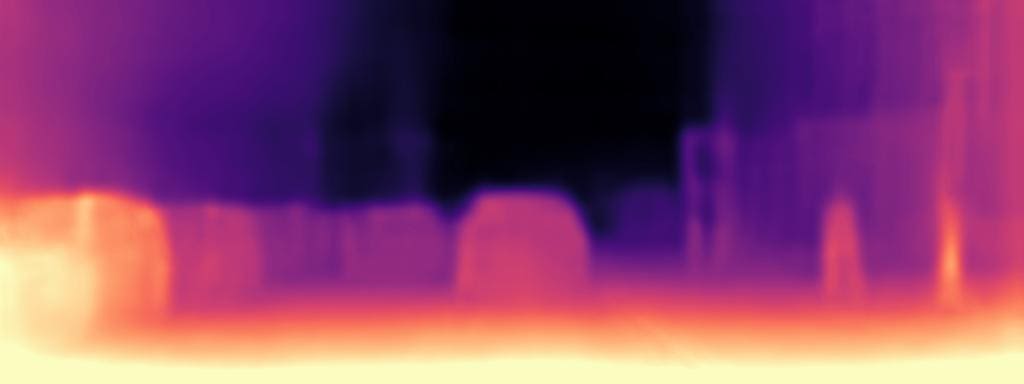}\hfill
\includegraphics[width=0.165\linewidth]{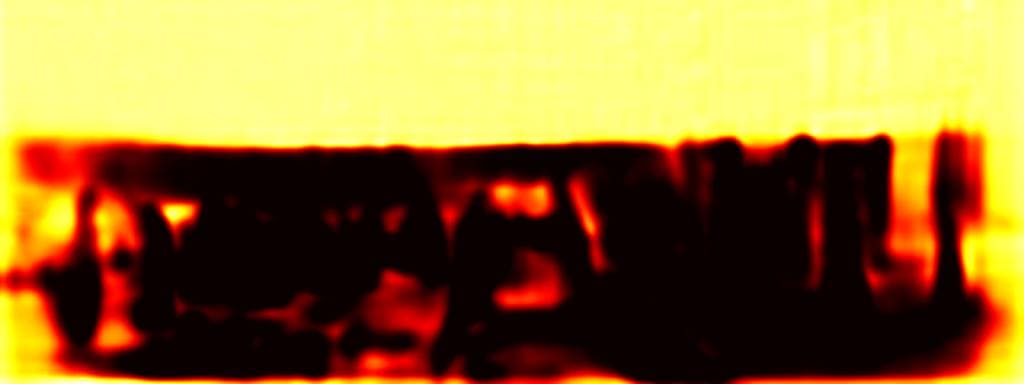}\hfill
\includegraphics[width=0.165\linewidth]{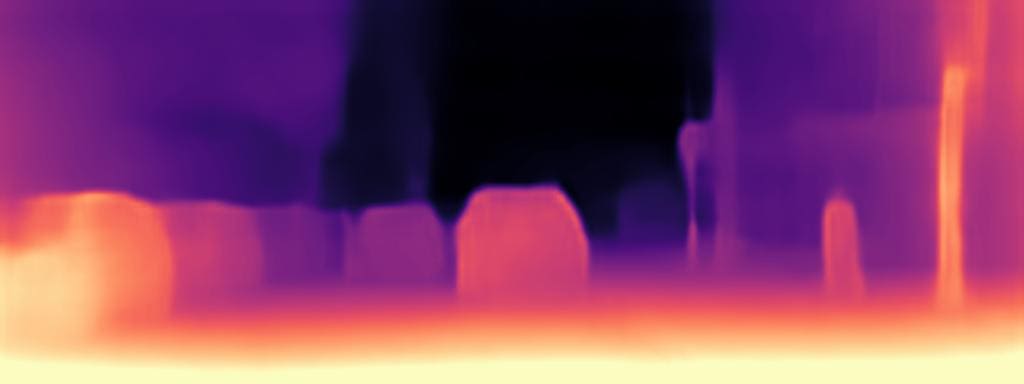}\hfill
\includegraphics[width=0.165\linewidth]{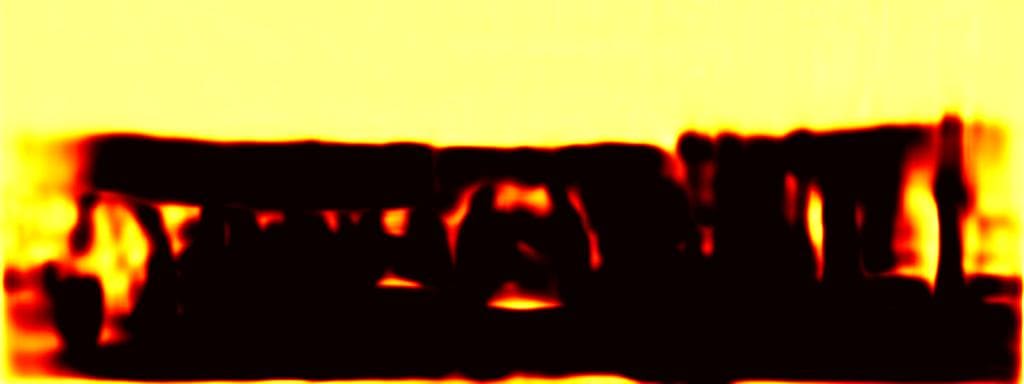}\hfill
\\ \vspace{+2pt}
\includegraphics[width=0.165\linewidth]{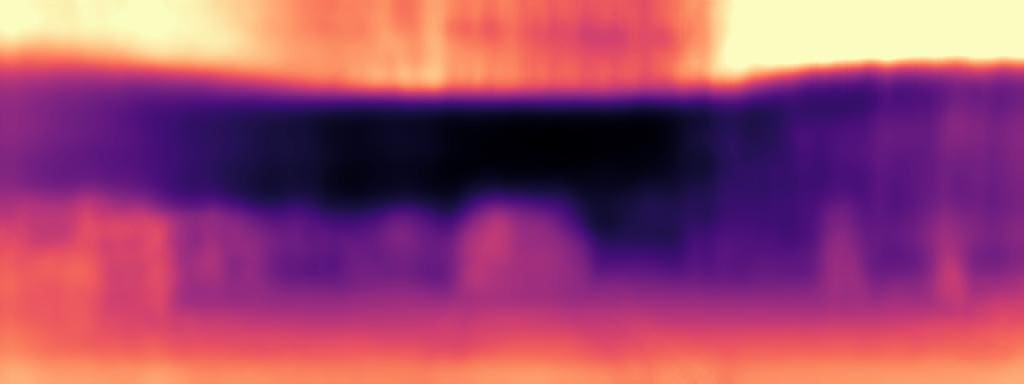}\hfill
\includegraphics[width=0.165\linewidth]{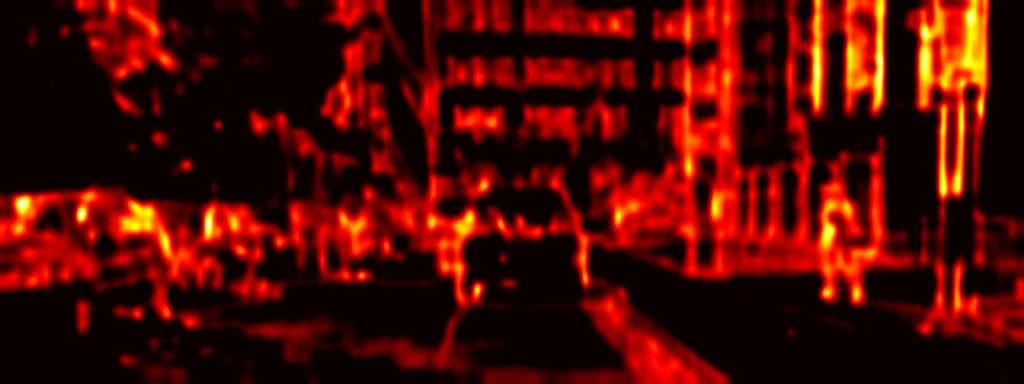}\hfill
\includegraphics[width=0.165\linewidth]{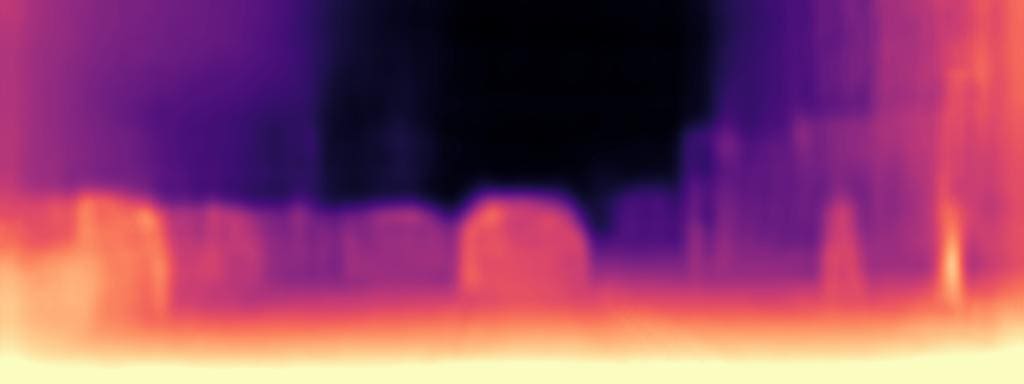}\hfill
\includegraphics[width=0.165\linewidth]{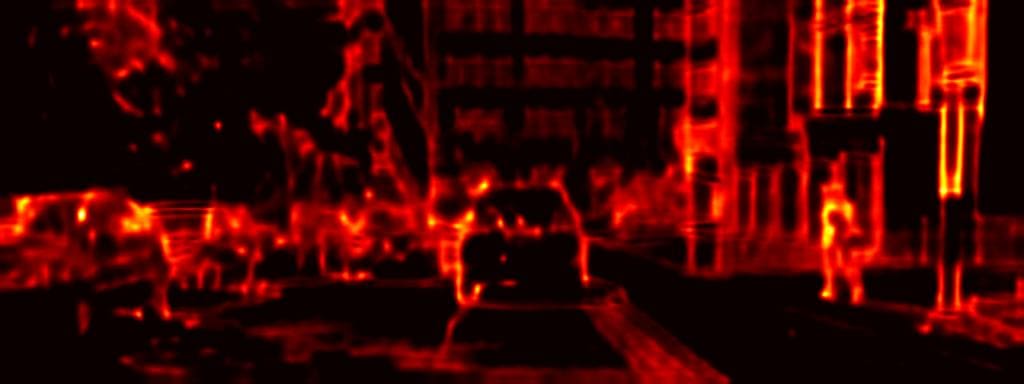}\hfill
\includegraphics[width=0.165\linewidth]{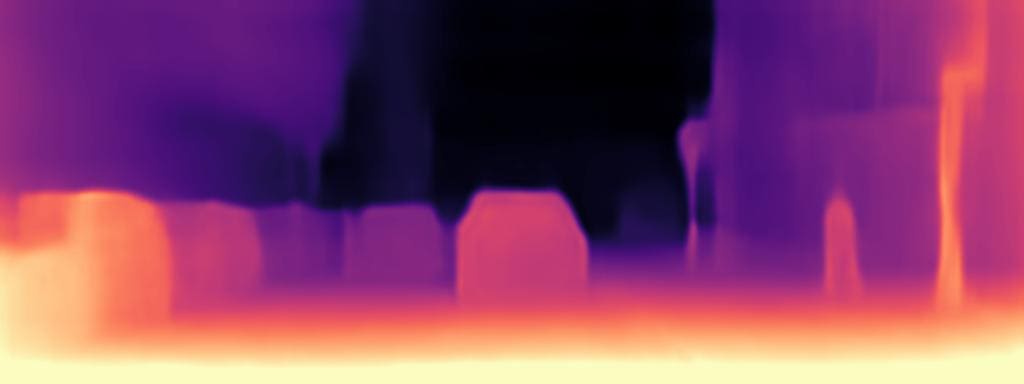}\hfill
\includegraphics[width=0.165\linewidth]{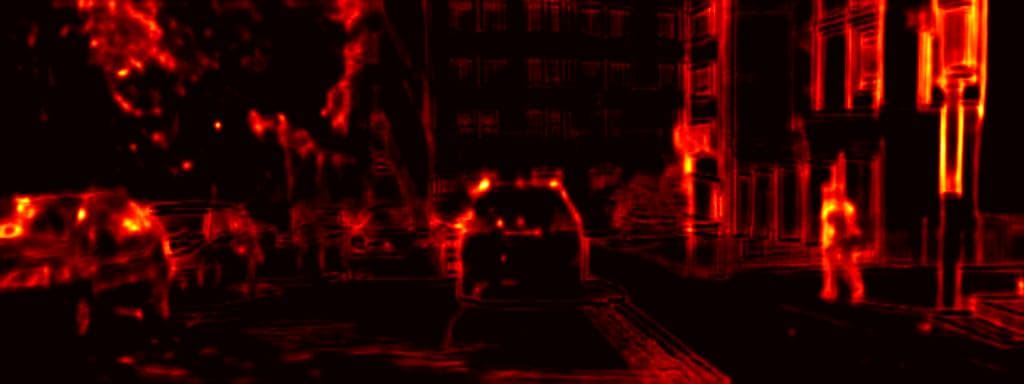}\hfill
\\ 
  \caption{\textbf{Convergence of our framework:} depths and corresponding uncertainties by (top) \textit{supervised} branch and (bottom) \textit{unsupervised} branch within our framework. As evolving iterations, each network generates better depths and uncertainties, which provides complementary information, and thus mutually boosts each other within our framework.}
\label{iteration}\vspace{-10pt}
\end{figure*}

Instead of relying on the unsupervised loss function, some methods~\cite{guo2018learning,cho2019large,tonioni2019unsupervised} attempted to train the monocular depth estimation network through pseudo depth by stereo matching, as illustrated in~\figref{fig_2}(b). To mitigate performance degeneration by inaccurate pseudo depth, they leverage additional confidence estimation networks~\cite{poggi2016learning,tosi2018beyond} to measure the confidence $C$ of pseudo depth $D_\mathrm{PGT}$. The monocular depth estimation networks then learn such stereo knowledge through a distillation loss as
\begin{equation}
    L = \frac{1}{N_C}\sum_{i\in\Omega} C'(i) \|D(i)-D_\mathrm{PGT}(i)\|_1,
\end{equation}
where $N_C$ is the number of pixels that $C'(i)=1$, and $C'$ is thresholded from $C$. Although the performance improvement was apparent~\cite{guo2018learning,cho2019large,tonioni2019unsupervised}, demanding additional stereo matching modules, which are often complex to measure the matching costs across disparities, hinders their applicability. Using additional confidence estimation networks is another computational burden~\cite{poggi2016learning}. 

To overcome the aforementioned limitations of conventional methods~\cite{kuznietsov2017semi,cho2019large}, we present a novel 
\textit{semi-supervised} learning framework for learning monocular depth estimation networks. 
To achieve the complementary advantages of both sparse supervised loss and unsupervised loss, as illustrated in~\figref{fig_3}, we design two independent networks with same architecture tailored for each loss function, called \textit{supervised branch} and \textit{unsupervised branch}. They are learned independently and distiled each other through the proposed \textit{mutual distillation} loss function. To extract confident pseudo depth labels from each network, we learn the networks in a probabilistic fashion to estimate the distribution of output depth. The uncertainties are used to adjusting the reliability of the depth maps when distilling the depth knowledge of one network to the other network. We also apply two different data augmentation to two branches.

\subsection{Mutual Distillation for Semi-Supervised Learning}
In this section, we describe the loss functions defined for each branch and mutual distillation loss.

\textbf{Learning Depth and Uncertainty.}
Considering the uncertainty and ignoring the regions with high uncertainty enable transferring reliable depth knowledge to each other in our framework. To this end, we leverage a negative log-likelihood minimization to infer the uncertainty, as well as depth, as in~\cite{poggi2020uncertainty}. Specifically, the predictive distribution of the output $D$ can be modelled as the Laplacian likelihood~\cite{kendall2017uncertainties} as
\begin{equation}
    L'_\mathrm{S,u} = 
    \frac{1}{N_D}\sum_{i\in\Omega_D} 
    \left(\frac{\|D_\mathrm{S}(i)-D_\mathrm{GT}(i)\|_1}{\sigma_\mathrm{S}(i)} + \mu_\mathrm{S}\mathrm{log}(\sigma_\mathrm{S}(i))\right),
\end{equation}
where $D_\mathrm{S}$ and $\sigma_\mathrm{S}$ denote the predicted depth and its corresponding uncertainty, respectively. $\mu_\mathrm{S}$ is a hyperparameter. The additional logarithmic term prevents the uncertainty from approaching infinite predictions. However, due to the sparsity nature of ground-truth depth maps, minimizing the above loss function cannot recover all the pixels in an image. For instance, the estimated depth maps and uncertainty maps would be ambiguous at the sky or upper parts of an image, out of the field of view of LiDAR, as the networks have never seen the ground-truth depths at such regions. Even though the depth quality is poor at the regions, if the uncertainty estimation is reliably measured, such unreliable depth labels could be ignored when transferring depth knowledge. To overcome this, we introduce additional loss term to deal with such regions, by indicating such regions unreliable, and the loss function is then reformulated such that
\begin{equation}
    L_\mathrm{S,u} = 
    L'_\mathrm{S,u} +
    \frac{1}{N_{D/}} \sum_{j\in\Omega_{D/}} 
    \left(\frac{M}{\sigma_\mathrm{S}(j)}+\mu_\mathrm{S}\mathrm{log}(\sigma_\mathrm{S}(j))\right),
\end{equation}
where $\Omega_{D/}$ denotes the pixels outside $\Omega_{D}$, and $M$ is a hyperparameter. The latter term is called \textit{unprojected point filtering} loss. \figref{unprojected} visualizes the effect of the term. Note that our loss function $L_\mathrm{S,u}$ is the first attempt to simultaneously predict the depth and its uncertainty under sparse depth supervisions.

We similarly learn the unsupervised branch with the loss:
\begin{equation}
    L_\mathrm{U,u} = \frac{1}{N} \sum_{i\in\Omega}
    \left(\frac{\mathrm{pe}(I(i),I'(i))}{\sigma_\mathrm{U}(i)} + \mu_\mathrm{U}\mathrm{log}(\sigma_\mathrm{U}(i))\right).
\end{equation}
\begin{table*}
\begin{center}
\scalebox{0.95}{
\begin{tabular}{l|c|c|c|cccc|ccc} 
\hlinewd{0.8pt}
\multirow{2}{*}{Methods} & \multirow{2}{*}{Supervision} 
&\multirow{2}{*}{\# param.} 
&\multirow{2}{*}{time} 
&Abs Rel
&Sqr Rel
&RMSE 
&RMSE log 
&$\delta$ < 1.25 & $\delta^{2}$ < 1.25 & $\delta^{3}$ < 1.25       \\ 
\cline{5-11}
& & & & \multicolumn{4}{c|}{lower is better} & \multicolumn{3}{c}{higher is better}\\ 
\hline
Eigen et al.~\cite{eigen2014depth} & Sup &54M& - &0.203&\underline{1.548}&\underline{6.307}&0.282& \underline{0.702}&\underline{0.890}&0.890 \\
Liu et al.~\cite{liu2015learning} & Sup &40M &- &\underline{0.201}&1.584&6.471&\underline{0.273}&0.680&0.898&\underline{0.967} \\
\textbf{Ours} & Sup  & 14M & 2.9ms &\textbf{0.105}& \textbf{0.695} &\textbf{4.398}&\textbf{0.179} &\textbf{0.885}&\textbf{0.964}&\textbf{0.984} \\
\hline
Monodepth~\cite{godard2017unsupervised} & Self (S) & 56M & 9.4ms & 0.138& 1.186& 5.650 & 0.234& 0.813& 0.930& 0.969 \\
MonoResMatch~\cite{tosi2019learning}&Self (S)  &41M &8.3ms & 0.111&0.867&4.714&0.199& 0.864&0.954&0.979\\
Uncertainty~\cite{poggi2020uncertainty}& Self (S) &14M&3.6ms & 0.107& 0.811& 4.796& 0.200& 0.866& 0.952& 0.978  \\
PackNet-sfM~\cite{guizilini20203d}& Self (M) &122M &9.5ms & 0.111& 0.785& \textbf{4.601}& \textbf{0.189}& \underline{0.878}& \textbf{0.960}& \textbf{0.982} \\
DepthHint~\cite{watson2019self} & Self (S) &33M &6.6ms & \textbf{0.102}& \textbf{0.762}& \underline{4.602}& \textbf{0.189}&\textbf{0.880} & \textbf{0.960}& \underline{0.981}\\
Insta-DM~\cite{lee2021learning} & Self (S) &14M &3.0ms & 0.112& \underline{0.777}& 4.772& \underline{0.191}& 0.872 & \underline{0.959}& \textbf{0.982}\\
\textbf{Ours  (Monodepth2~\cite{godard2019digging})} & Self (MS) &14M &2.9ms & \underline{0.106}& 0.818&  4.750& 0.196& 0.874& 0.957 & 0.979\\
\hline
SVSM FT~\cite{mayer2016large}& Semi (S)  & -& - & 0.102& 0.700& 4.681& 0.200& 0.872& 0.954& 0.978  \\
Kuznietsov et al.~\cite{kuznietsov2017semi}  & Semi (S)  &81M &   -& 0.113& 0.741& 4.621& 0.189& 0.862& 0.960& \textbf{0.986} \\
OnboardDepth.~\cite{angelova2019onboarddepth}& Semi (S) & -  &     - & 0.115& 0.766& 4.665& 0.189& 0.861& 0.957& 0.983     \\
Cho et al.~\cite{cho2019large}& Semi (S) & -  &     - & \textbf{0.099}& 0.748& 4.599& \underline{0.183}& \underline{0.880}& 0.959& 0.983     \\

\textbf{Ours}    & Semi (S)     & 18M &  2.9ms  & 0.117 &0.753& 4.423& 0.273&  0.870& 0.946& 0.965 \\
\textbf{Ours}    & Semi (M)     & 18M&   2.9ms    & \underline{0.101}& \underline{0.673}& \underline{4.292}& \textbf{0.176}&\textbf{0.892}& \underline{0.965}& \underline{0.984}  \\
\textbf{Ours}    & Semi (MS) & 18M  &  2.9ms    & \underline{0.101}& \textbf{0.657}& \textbf{4.262}& \textbf{0.176}&  \textbf{0.892}& \textbf{0.966}& \underline{0.984} \\
\hlinewd{0.8pt}
 \end{tabular}}
\end{center}\vspace{-5pt}
\caption{\textbf{Quantitative results on KITTI dataset~\cite{geiger2012we}.} The best results are in \textbf{bold}, and the second best results are \underline{underlined}. Self, Sup and Semi respectively indicate self-supervised, supervised and semi-supervised learning, with (M), (S) and (MS) respectively indicating monocular, stereo, and both. Our method trained with self-supervised learning on monocular stereo videos is exactly same as Monodepth2~\cite{godard2019digging} (denoted Ours (Monodepth2~\cite{godard2019digging}).}\label{tab:evalresults}\vspace{-5pt}
\end{table*}
\begin{figure*}[t]
	\centering
    \captionsetup[subfigure]{labelformat=empty}
	\subfloat[]
	{\includegraphics[width=0.165\linewidth]{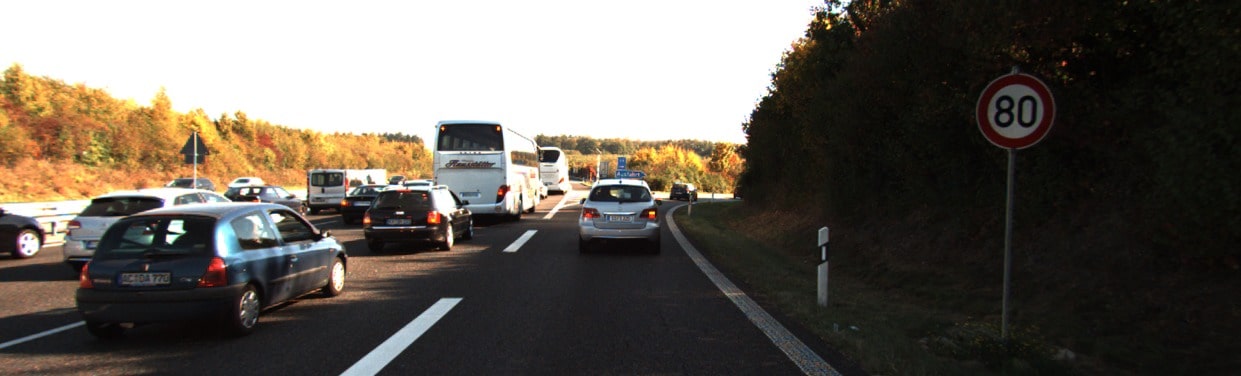}}\hfill
    \subfloat[]
	{\includegraphics[width=0.165\linewidth]{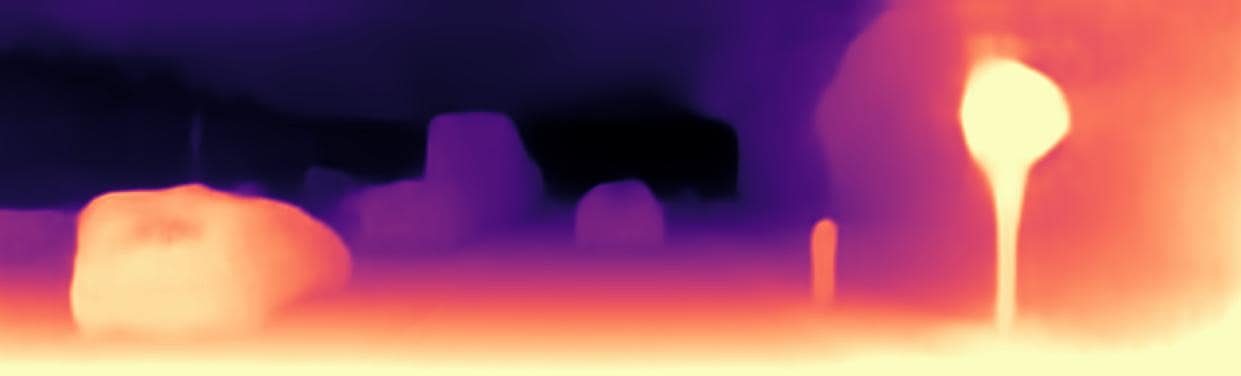}}\hfill
	\subfloat[]
	{\includegraphics[width=0.165\linewidth]{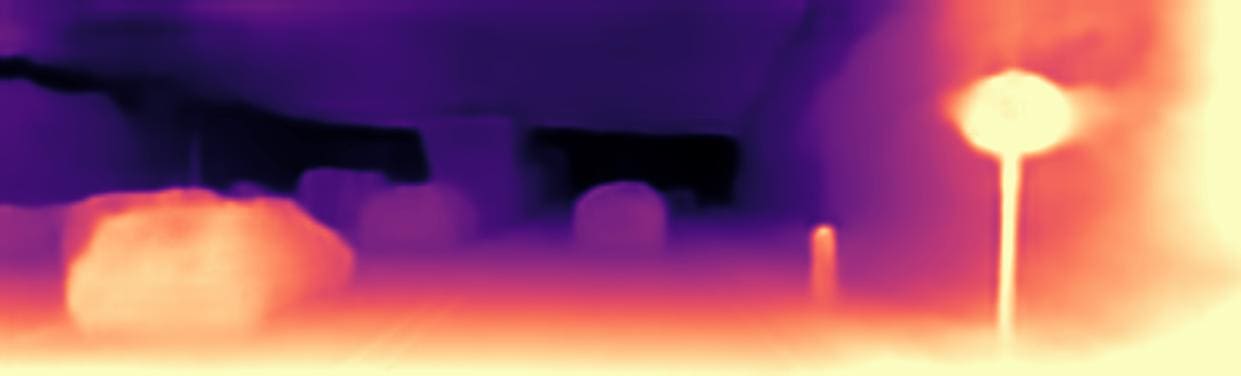}}\hfill
    \subfloat[]
      	{\includegraphics[width=0.165\linewidth]{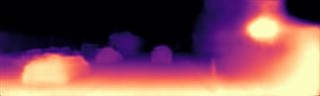}}\hfill
	\subfloat[]
	{\includegraphics[width=0.165\linewidth]{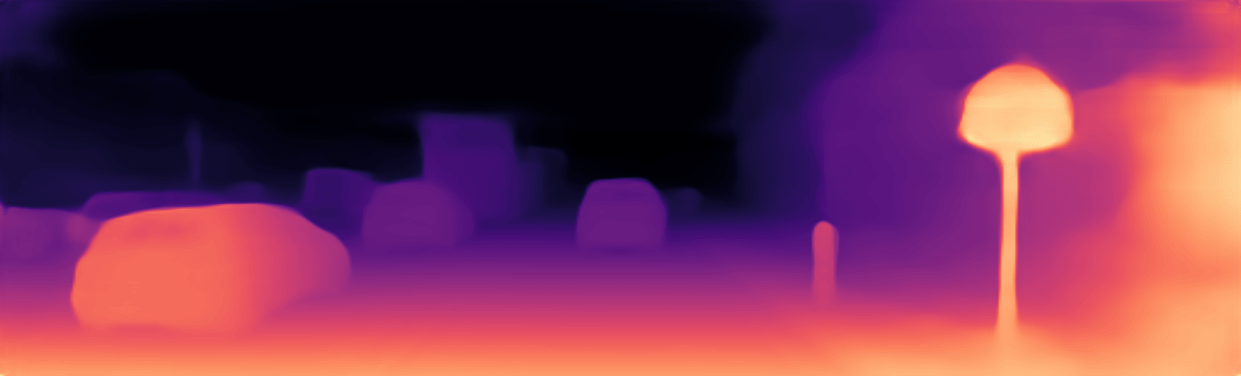}}\hfill
	\subfloat[]	
	{\includegraphics[width=0.165\linewidth]{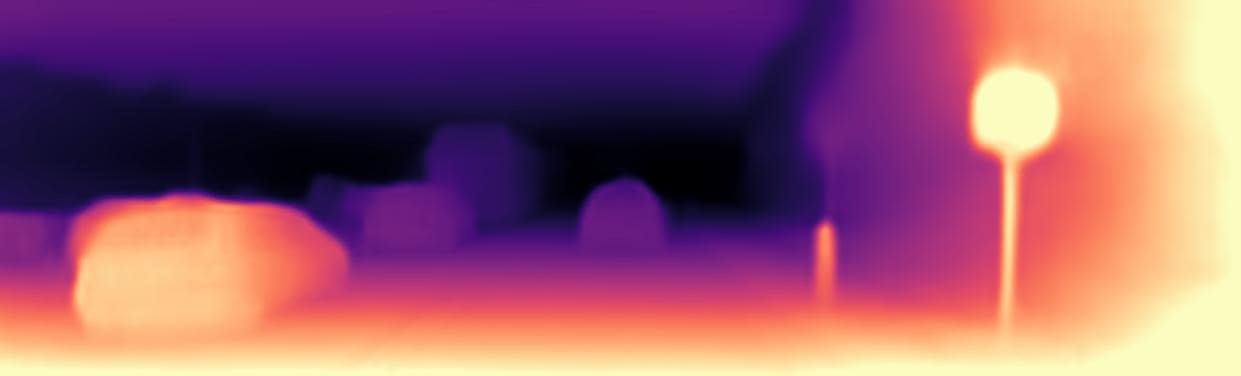}}\hfill\\\vspace{-20.5pt}
	\subfloat[]
	{\includegraphics[width=0.165\linewidth]{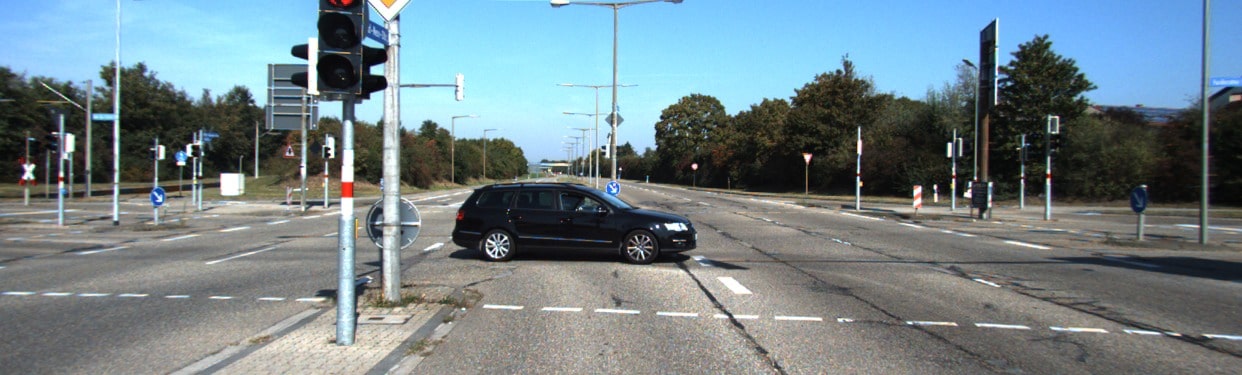}}\hfill
    \subfloat[]
	{\includegraphics[width=0.165\linewidth]{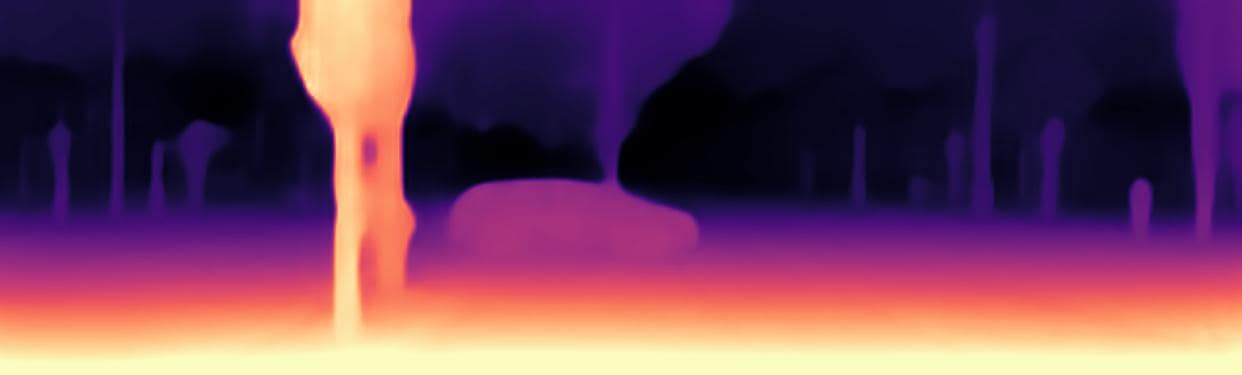}}\hfill
	\subfloat[]
	{\includegraphics[width=0.165\linewidth]{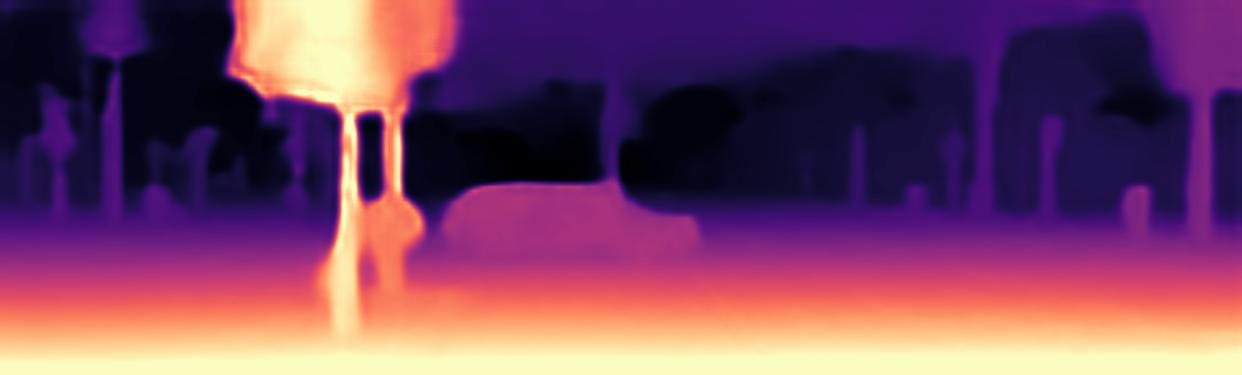}}\hfill
    \subfloat[]
	{\includegraphics[width=0.165\linewidth]{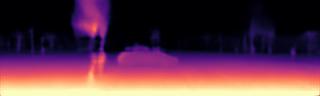}}\hfill
	\subfloat[]
	{\includegraphics[width=0.165\linewidth]{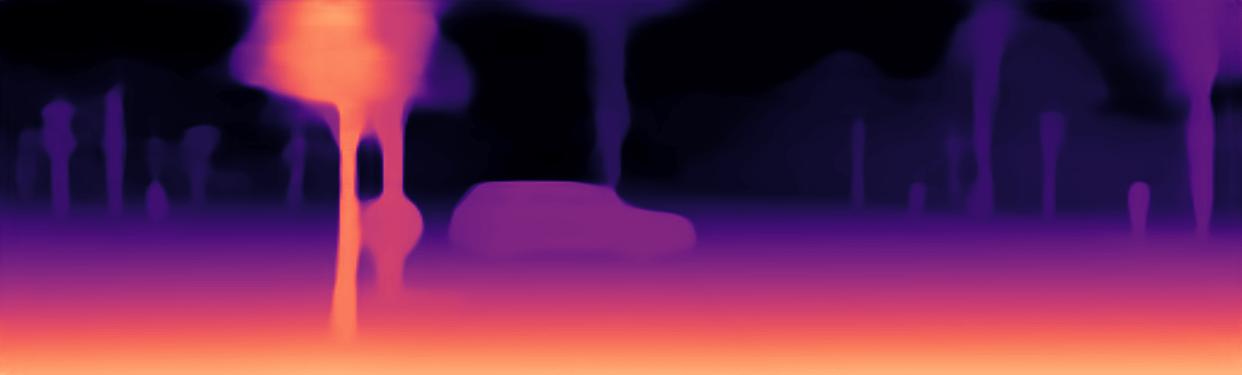}}\hfill
	\subfloat[]	
	{\includegraphics[width=0.165\linewidth]{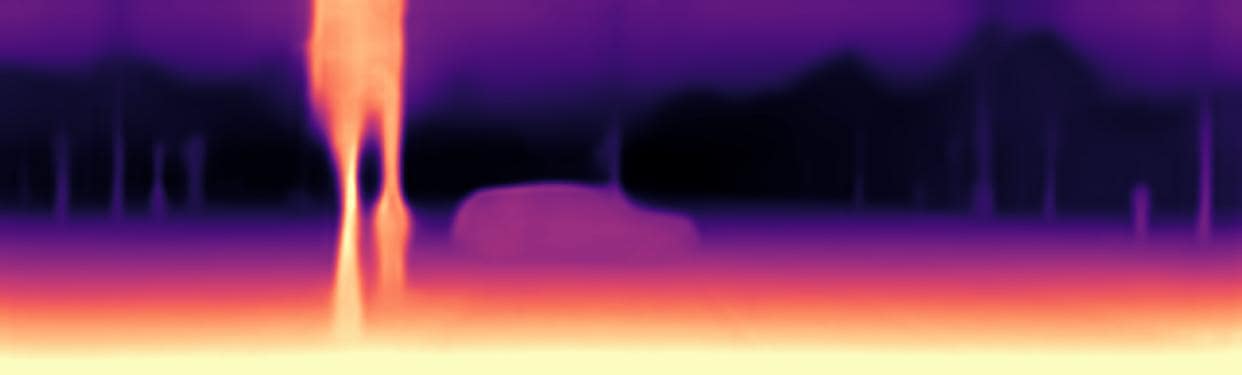}}\hfill\\\vspace{-20.5pt}
	\subfloat[]
	{\includegraphics[width=0.165\linewidth]{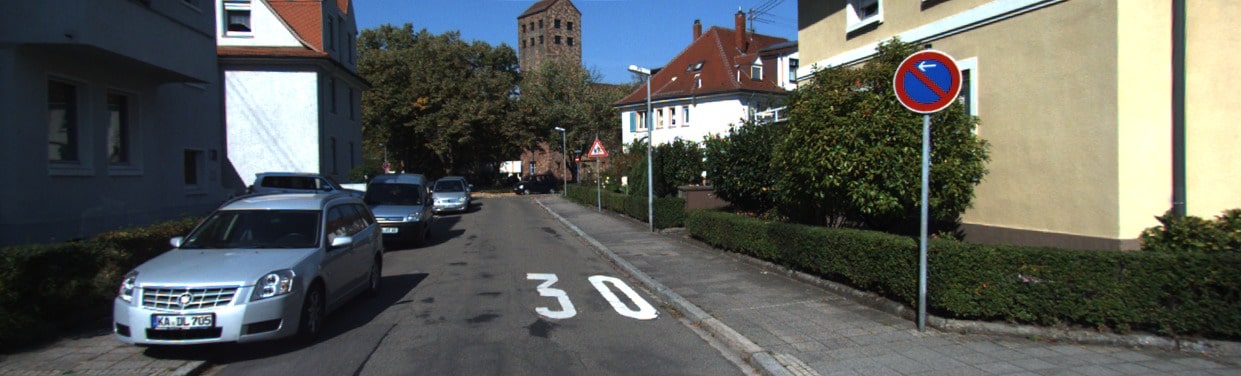}}\hfill
    \subfloat[]
	{\includegraphics[width=0.165\linewidth]{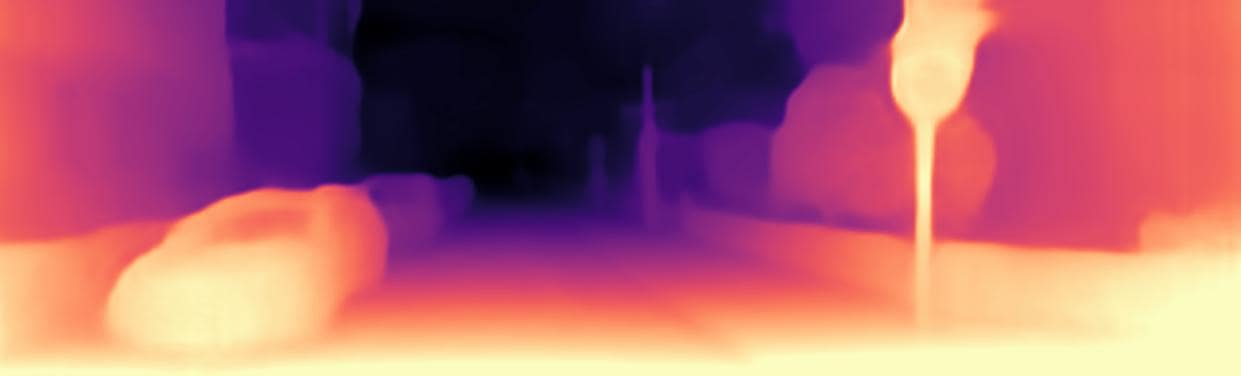}}\hfill
	\subfloat[]	
	{\includegraphics[width=0.165\linewidth]{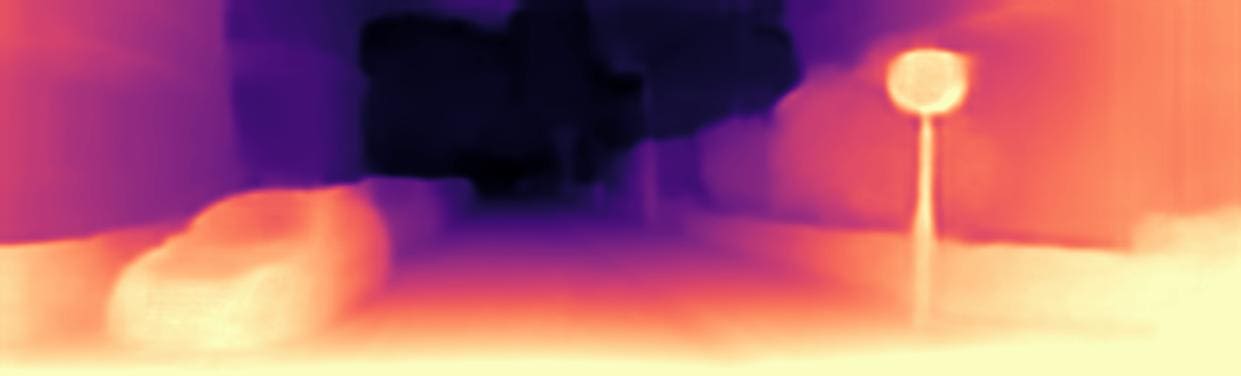}}\hfill
	\subfloat[]	
	{\includegraphics[width=0.165\linewidth]{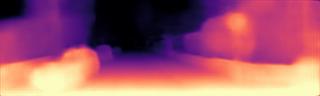}}\hfill
	\subfloat[]	
	{\includegraphics[width=0.165\linewidth]{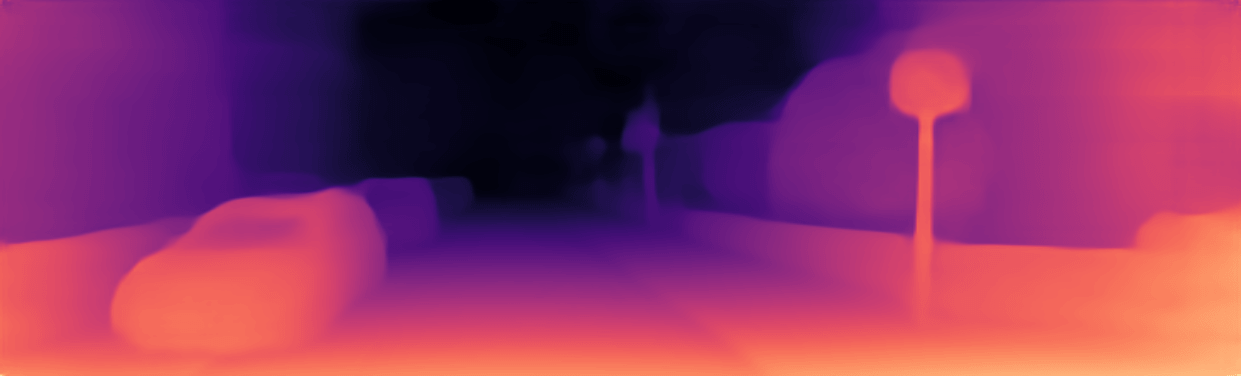}}\hfill
	\subfloat[]	
	{\includegraphics[width=0.165\linewidth]{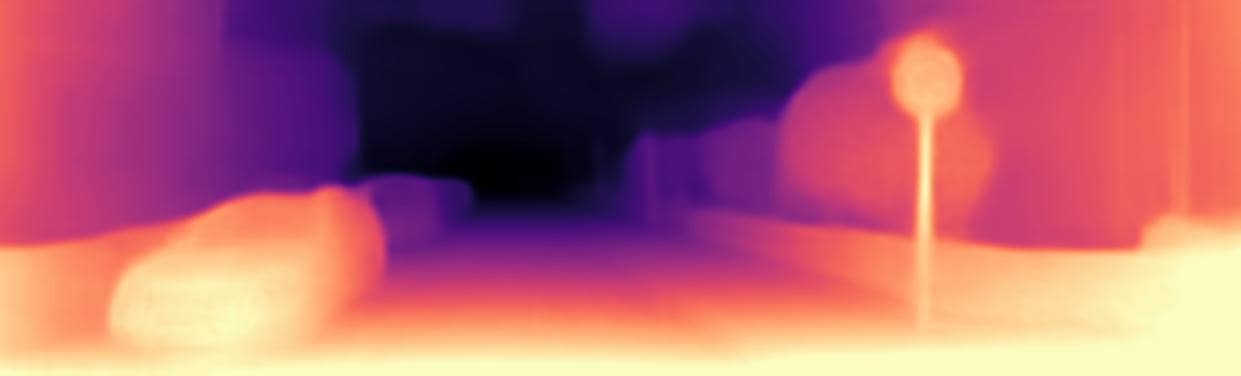}}\hfill\\
	\vspace{-20.5pt}
	\subfloat[(a) Color image]
	{\includegraphics[width=0.165\linewidth]{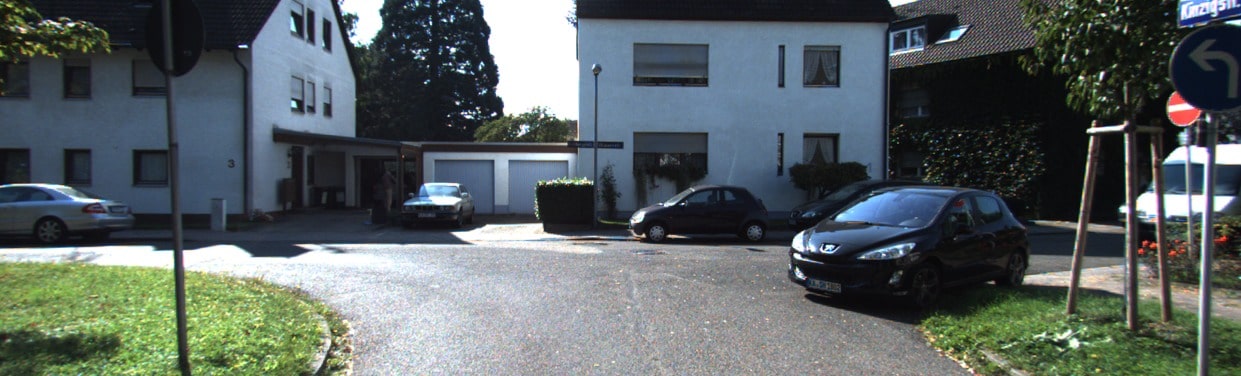}}\hfill
    \subfloat[(b) Monodepth2~\cite{godard2019digging}]
	{\includegraphics[width=0.165\linewidth]{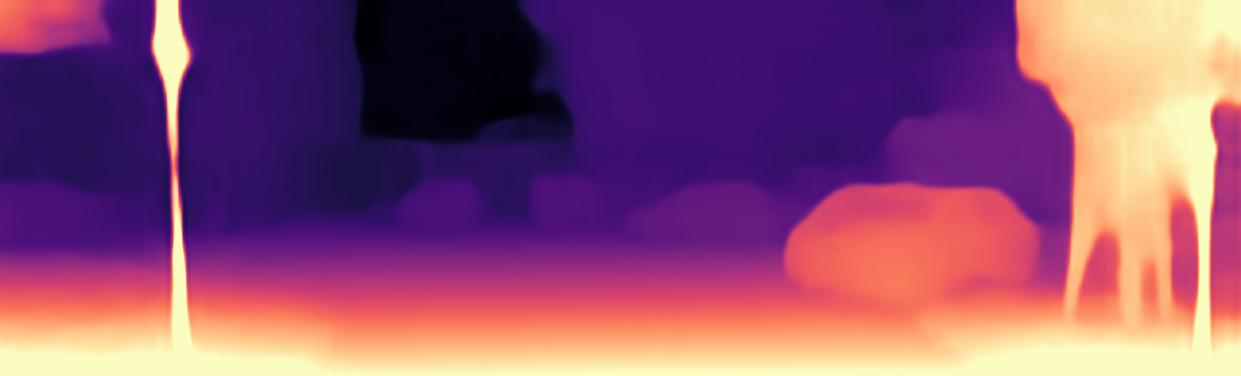}}\hfill
	\subfloat[(c) DepthHint~\cite{watson2019self}]
	{\includegraphics[width=0.165\linewidth]{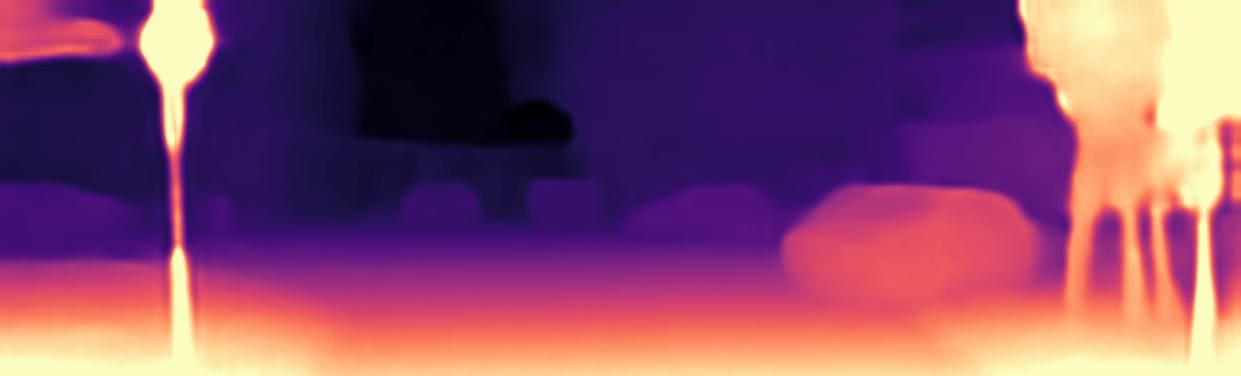}}\hfill
    \subfloat[(d) kuznietsov et al.~\cite{kuznietsov2017semi}]
	{\includegraphics[width=0.165\linewidth]{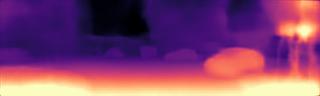}}\hfill
	\subfloat[(e) Amiri et al.~\cite{amiri2019semi}]
	{\includegraphics[width=0.165\linewidth]{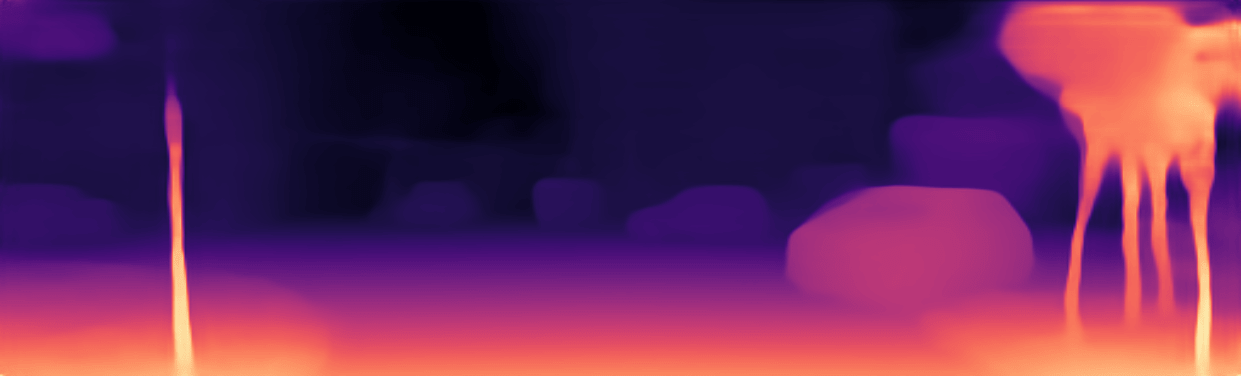}}\hfill
	\subfloat[(f) Ours]	
	{\includegraphics[width=0.165\linewidth]{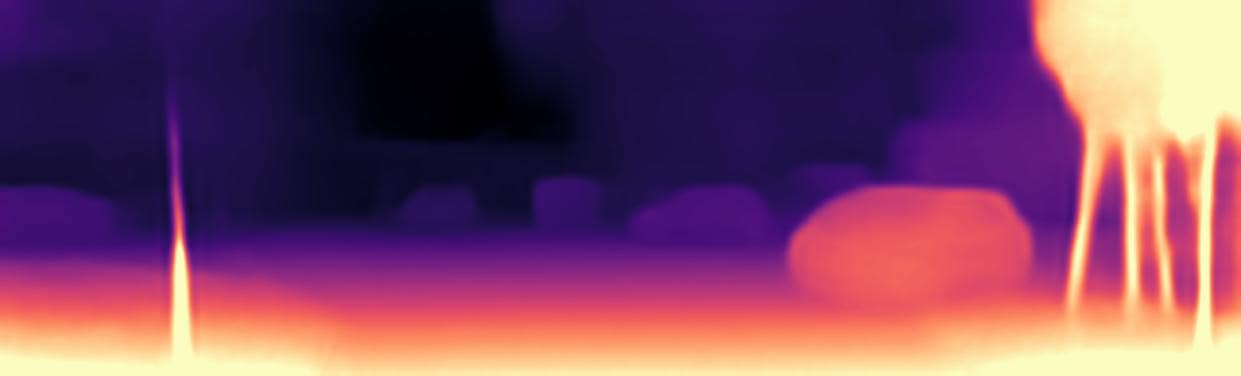}}\hfill\\
    \vspace{-5pt}
	\caption{\textbf{Qualitative results on KITTI datasets~\cite{geiger2012we}.} Comparing with existing methods, our model produces plausible depth maps better aligned with input images and recovers complex objects such as thin poles, trees, and traffic sign well.}
	\label{fig:kitti}\vspace{-10pt}
\end{figure*}

\textbf{Distillation of Depth with Uncertainty.}
By using the predicted depth map and its uncertainty map, we formulate a \textit{mutual distillation} loss function to mutually boost each network. Specifically, to transfer the confident depth knowledge from unsupervised branch to supervised branch, we define the distillation loss function $L_\mathrm{S,d}$ such that
\begin{equation}
    L_\mathrm{S,d} = \frac{1}{N_{\mathrm{U}}} \sum_{i\in\Omega} \frac{\|D_\mathrm{S}(i)-D_\mathrm{U}(i)\|_1}{\sigma_\mathrm{U}(i)},
\end{equation}
where $N_{\mathrm{U}} = \sum_i (1/\sigma_\mathrm{U}(i))$. The loss function $L_\mathrm{U,d}$ is similarly defined such that
\begin{equation}
    L_\mathrm{U,d} = \frac{1}{N_{\mathrm{S}}} \sum_{i\in\Omega} \frac{\|D_\mathrm{U}(i)-D_\mathrm{S}(i)\|_1}{\sigma_\mathrm{S}(i)}.
\end{equation}
Unlike conventional methods that leverage the stereo knowledge and its thresholded confidence~\cite{guo2018learning,tonioni2019unsupervised,cho2019large,watson2020learning}, we exploit the predicted uncertainty itself since it empirically yields better performance. In addition, unlike conventional semi-supervised methods~\cite{kuznietsov2017semi,amiri2019semi} that only leverage the fixed, sparse ground-truth depth maps as labeled data, our approach enlarges the pseudo labels as evolving the training, dramatically boosting the performance. \figref{iteration} visualizes the depth and confidence at two branches as evolving iterations, which shows complementary information of them.

\textbf{Total Loss.}
By considering the loss functions discussed so far, the total loss functions are defined such that: for supervised branch, $L_\mathrm{S,total} = L_\mathrm{S,u} + \lambda_\mathrm{S} \, L_\mathrm{S,d}$ and for unsupervised branch, $L_\mathrm{U,total} = L_\mathrm{U,u} + \lambda_\mathrm{U} \, L_\mathrm{U,d} + \lambda_\mathrm{smooth} \, L_\mathrm{smooth}$, where $\lambda_\mathrm{S}$, $\lambda_\mathrm{U}$, and $\lambda_\mathrm{smooth}$ are weighting parameters.


\subsection{Noising the Networks: Data Augmentation}
In many literature for semi-supervised learning~\cite{rawat2017deep,yalniz2019billion,xie2020self}, some methods attempted to learn robust features by giving different perturbations to separate networks and aligning the features. Inspired by these, since our network consists of two separate branches, different kinds of data augmentation can be applied to improve the robustness. To improve the ability to recover an object instance without relying on the bias from the background context or geometric structure of background such as the vanishing point, we present a novel augmentation technique for monocular depth estimation that injects a photometric noise which is utilized in \cite{godard2019digging} into an image except for object instances to help predict an instance-aware depth.

\section{Experimental Results}
\subsection{Implementation Details}
We implement our networks with the Pytorch library~\cite{paszke2017automatic}. Our monocular depth estimation network is based on U-net architecture~\cite{ronneberger2015u}, similar to Monodepth2~\cite{godard2019digging}, in which one \textit{shared} encoder extracts low resolution, high-dimensional features from the input image $I$ and two \textit{separate} decoders estimate the depth $D$ and its uncertainty $\sigma$, respectively. We design the decoder as a mirrored version of the encoder.

We conduct all our experiments with 24GB RTX-3090 GPU. We set the learning rate as ${10^{-4}}$ and batches of images downsampled to 640 × 192 as 16. We use Adam optimizer with $\beta_{1} = 0.9$ and $\beta_{2} = 0.999$. We set the SSIM weight as $\alpha=0.85$, following~\cite{godard2017unsupervised}. We set the weight parameters such that $\lambda_\mathrm{S}=1$, $\lambda_\mathrm{U}=0.05$, $L_\mathrm{smooth}=0.001$, $\mu_\mathrm{S}=3$, and $\mu_\mathrm{U}=0.03$, determined by cross-validation. We use the proposed data augmentation, as well as flipping and jittering, widely used in many literature~\cite{dwibedi2017cut,ghiasi2021simple}. For uncertainty estimation, we train the network to predict the log variance $\mathrm{log}(\sigma)$ because it is more numerically
stable than regressing the variance as the loss avoids any division by zero. Similar to Monodepth2~\cite{godard2019digging}, we use multi-scale loss functions. We
will make our code publicly available in case of acceptance. 
\begin{table}
\centering
\scalebox{0.9}{
\begin{tabular}{l|cccc|c} 
\hlinewd{0.8pt}
Methods & Abs & Sq & RMSE & RMSElog    & $\delta$ < 1.25  \\ 
\hline
Monodepth~\cite{godard2017unsupervised}  & 0.631 & 10.257 &13.424 &0.525 &0.281  \\
MonoResMatch~\cite{tosi2019learning}     & \underline{0.241} & 2.149  &9.064  &0.296 &0.570  \\
PackNet-SfM~\cite{guizilini20203d}         & 0.245 & 2.240  &8.920  &0.298 &0.557  \\
Monodepth2~\cite{godard2019digging}       & 0.242 & 2.308  &8.563  &0.290 &0.591  \\
DepthHint~\cite{watson2019self}           & \textbf{0.220} & \underline{2.008}  &\underline{8.363}  &\underline{0.273} &\textbf{0.613}  \\
\textbf{Ours}                        & \textbf{0.220} & \textbf{1.955}  &\textbf{8.234}  &\textbf{0.270} &\underline{0.612}  \\
\hlinewd{0.8pt}
\end{tabular}}
\caption{\textbf{Quantitative results on Cityscape  dataset~\cite{cordts2016cityscapes} without fine-tune.}}\label{tab:cittyscape}\vspace{-10pt}
\end{table}

\subsection{Experimental Settings}
In this section, we conduct extensive evaluations to verify the robustness of our framework in comparison with existing methods such as MonoDepth~\cite{godard2017unsupervised}, Monodepth2~\cite{godard2019digging}, Uncertainty~\cite{poggi2020uncertainty}, MonoResMatch~\cite{tosi2019learning}, DepthHint~\cite{watson2019self}, PackNet-SfM~\cite{guizilini20203d}, Kuznietsov et al.~\cite{kuznietsov2017semi},  SVSM FT~\cite{mayer2016large}, and Amiri et al.~\cite{amiri2019semi}. 
In experiments, we use KITTI benchmark~\cite{geiger2012we} and Cityscape benchmark~\cite{cordts2016cityscapes}. We train the proposed networks on KITTI benchmark~\cite{geiger2012we} with pre-processing by the Zhou et al.~\cite{zhou2017unsupervised} to remove static frames, which results in 39,810 monocular triplets for training and 4,424 for validation. We use annotated depth maps, refined by~\cite{uhrig2017sparsity}, to train supervised networks. For Cityscapes, following~\cite{pilzer2018unsupervised}, we split stereo image pairs into 22,973 for training and 1,525 for test. We cropped the stereo images by discarding botton parts (the car hood) of 25\% and resized them. We evaluate our networks through the standard metrics, RMSE, RMSE log, absolute relative difference (Abs Rel), squared relative difference (Sq Rel), and $\delta$, presented in Eigen et al.~\cite{eigen2014depth}.

\subsection{Experimental Results}
\textbf{Results on KITTI.} We evaluated the monocular depth estimation performance on the KITTI Eigen Split~\cite{eigen2014depth}. \tabref{tab:evalresults} shows the comparison of our method with the state-of-the-art methods. Our approach achieves competitive performance in comparison to other methods. Regarding the model capacity, e.g., parameters or time, our method is highly comparable to methods that have bigger networks, e.g., PackNet~\cite{guizilini20203d}.
In addition, although our current model was based on unsupervised loss functions presented in Monodepth2~\cite{godard2019digging}, better unsupervised loss functions, as in DepthHint~\cite{watson2019self}, could boost the performance. In qualitative evaluations of~\figref{fig:kitti}, our model shows robust prediction results while preserving the shape of the object well compared to other methods.
\begin{figure}[t]
\centering
\subfloat[]
{\includegraphics[width=0.33\linewidth]{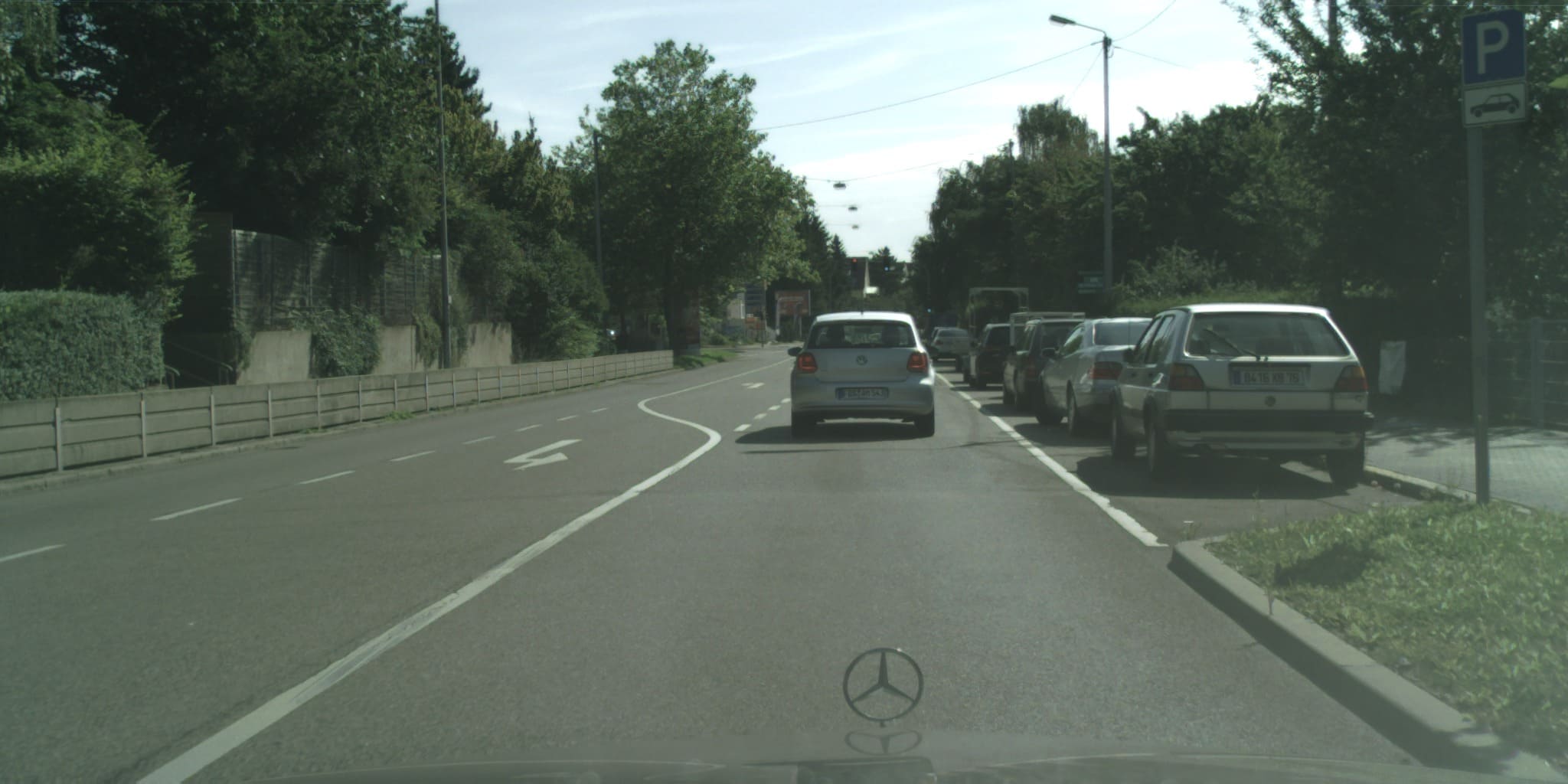}}\hfill
\subfloat[]
{\includegraphics[width=0.33\linewidth]{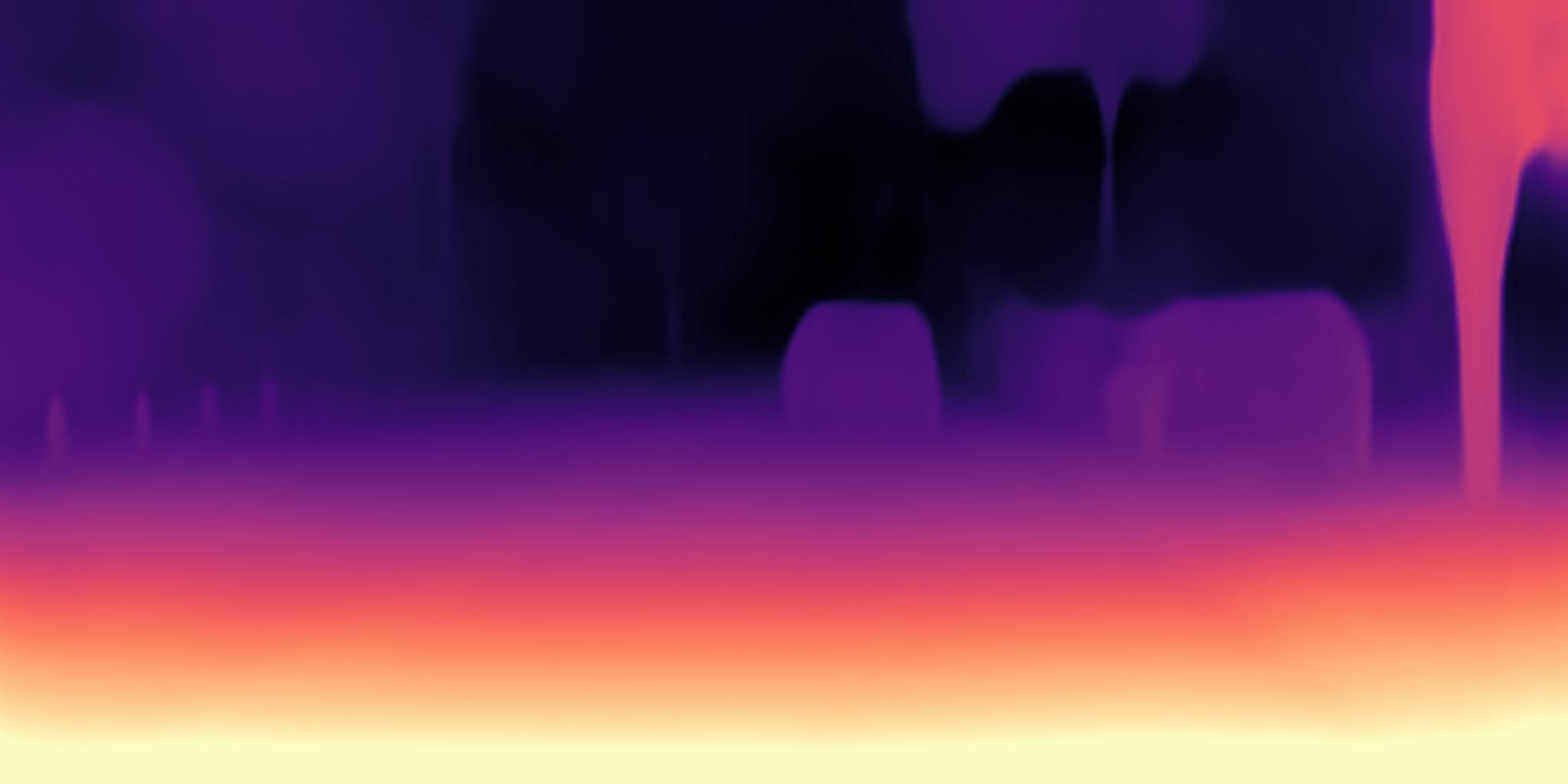}}\hfill
\subfloat[]
{\includegraphics[width=0.33\linewidth]{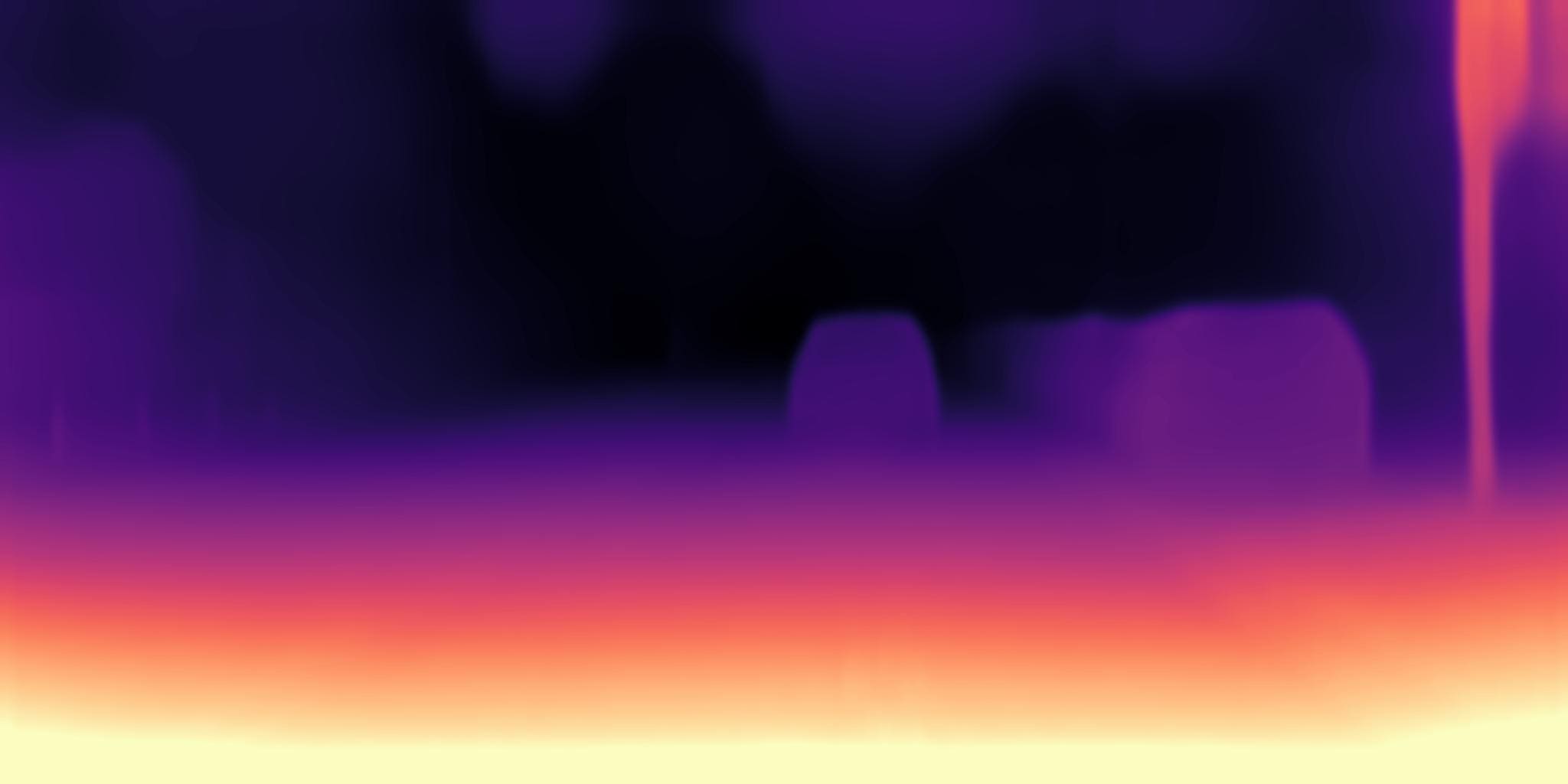}}\hfill
\\ \vspace{-20.5pt}
\captionsetup[subfigure]{labelformat=empty}
\subfloat[(a) Color]
{\includegraphics[width=0.33\linewidth]{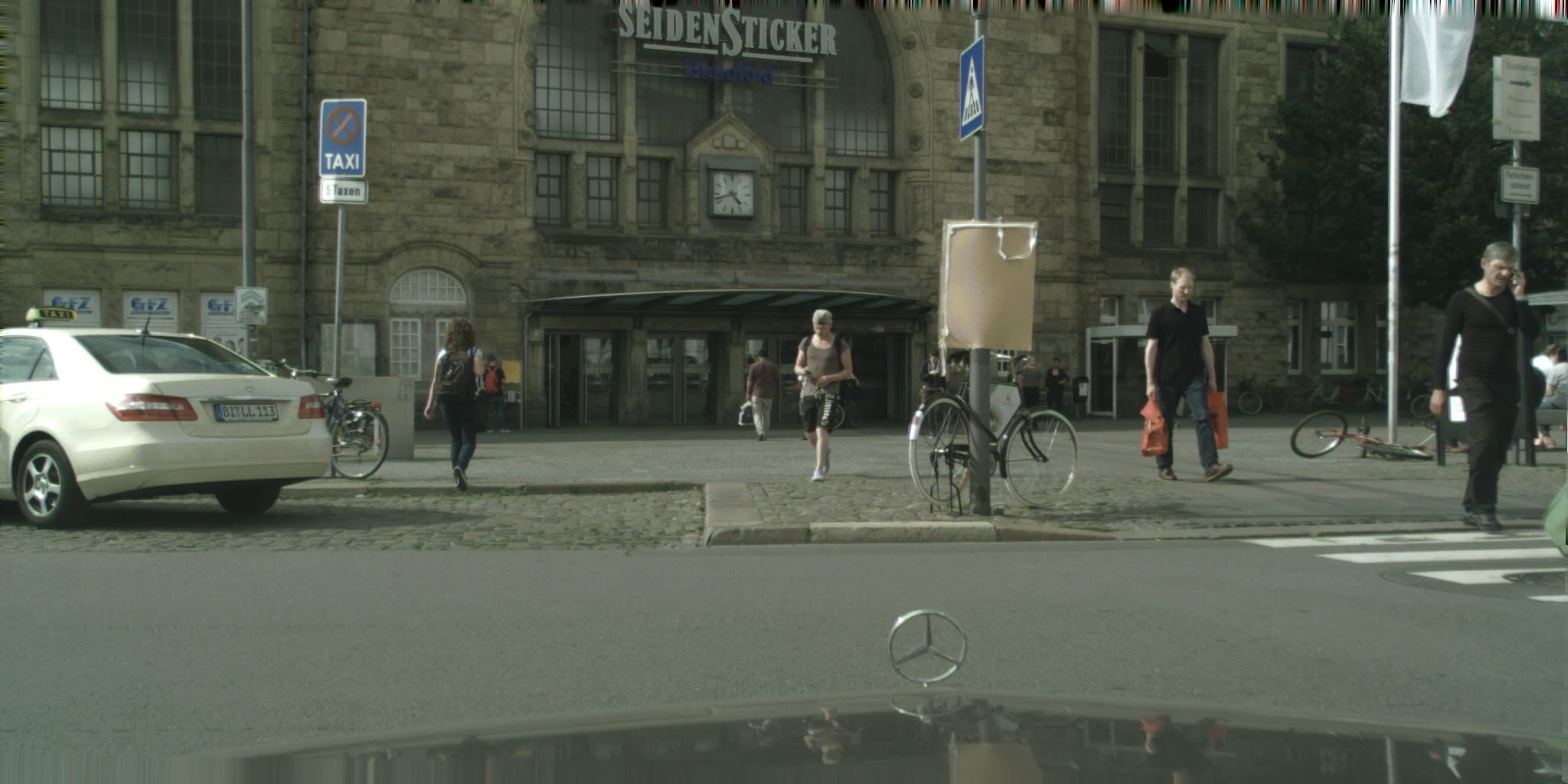}}\hfill
\subfloat[(b) Monodepth2~\cite{godard2019digging}]
{\includegraphics[width=0.33\linewidth]{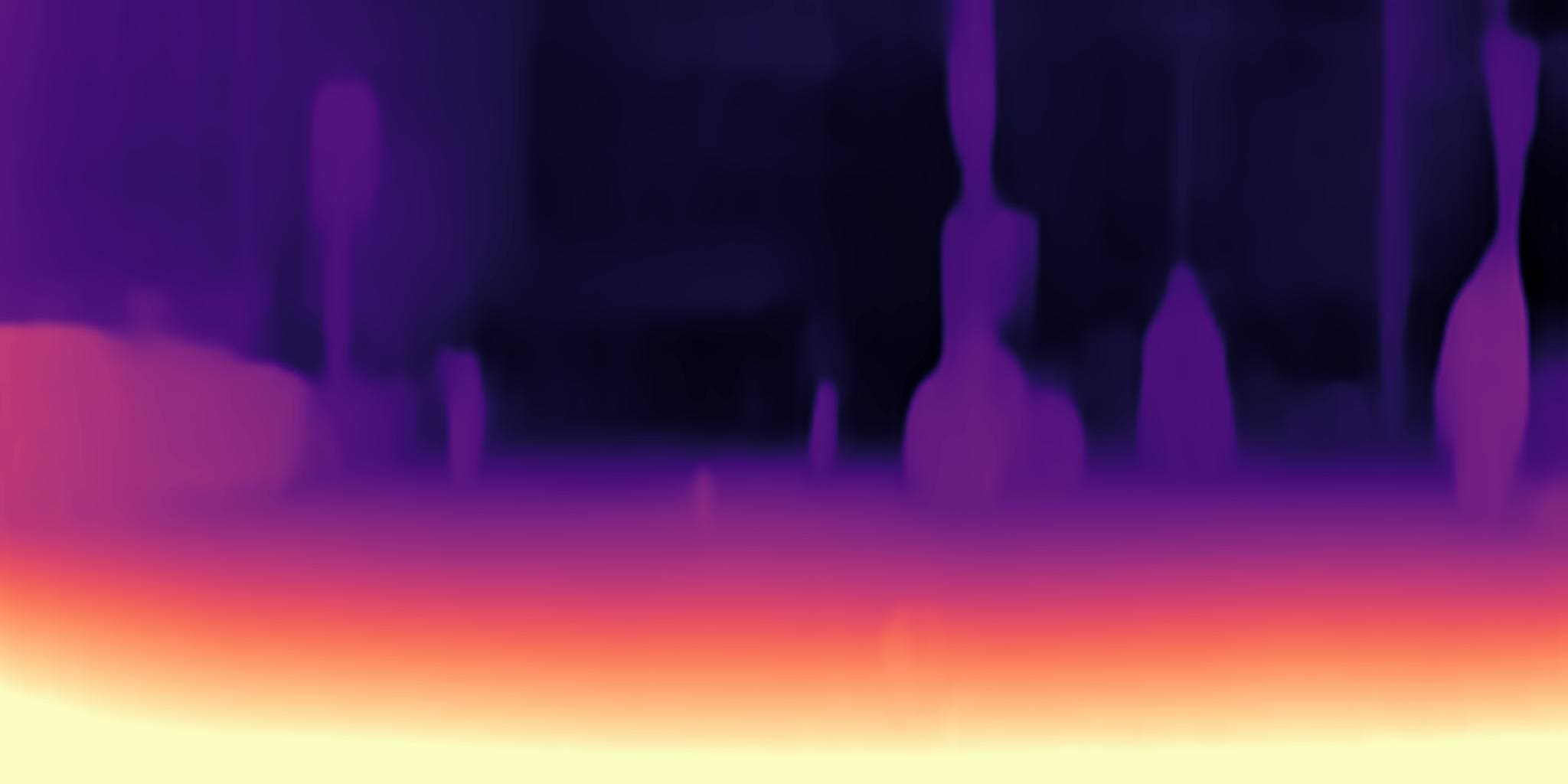}}\hfill
\subfloat[(c) Ours]
{\includegraphics[width=0.33\linewidth]{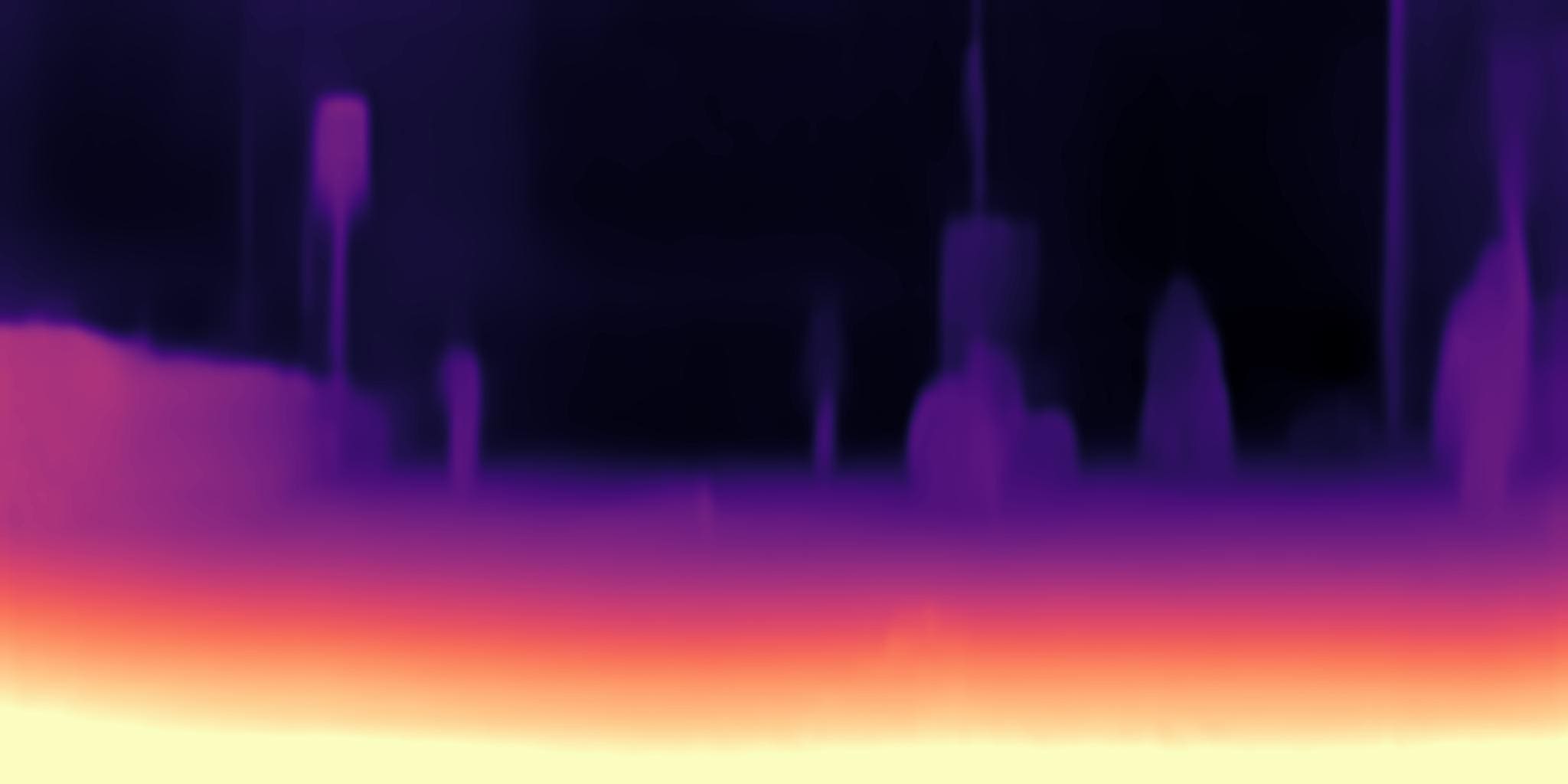}}\hfill
\\ \vspace{-5pt}
\caption{\textbf{Qualitative results on Cityscape dataset~\cite{cordts2016cityscapes}.}}
\label{city}\vspace{-5pt}
\end{figure}
\begin{table}[t]
\centering
\begin{tabular}{c|cccc|c} 
\hlinewd{0.8pt}
{Methods} & Abs & Sq & RMSE & RMSElog & $\delta$ < 1.25  \\ 
\hline
UW (Ours)      &\textbf{0.102}     &\textbf{0.664}     &\textbf{4.272}      &\textbf{0.179}       &\textbf{0.892}       \\
\hline
UT       &0.104     &0.669     &4.401      &0.180       &0.886      \\
UWT      &0.103     &0.665     &4.422      &0.180       &0.885       \\
CT       &0.107     &0.730     &4.301      &0.179       &0.887      \\ 
\hlinewd{0.8pt}
\end{tabular}
\caption{\textbf{Evaluation of mutual distillation strategy:} using uncertainty weighting (UW) (Ours), uncertainty thresholding (UT), both (UWT) and confidence thresholding (CT).}\label{tab:comparision}
\vspace{-10pt}
\end{table}
\begin{table}[t]
\centering
\begin{tabular}{l|ccc|cc|c} 
\hlinewd{0.8pt}
Methods                     & D            & M          & N         & Abs  & RMSE                 & $\delta$ < 1.25     \\ 
\hline
Baseline            &              &            &           &0.106 &4.750                 &0.874              \\ 
\hline 
Ours (D)                     &\checkmark    &            &           &0.103 &4.327                 &0.889                \\
Ours (D+M)                   &\checkmark    &\checkmark  &           &0.102 &4.272                 &0.892                \\
Ours (N)                     &              &            &\checkmark &0.107 &4.686                 &0.884                \\ 
Ours (D+M+N)           &\checkmark    &\checkmark  &\checkmark &\textbf{0.101} &\textbf{4.262} &\textbf{0.892}           \\ 
\hlinewd{0.8pt}
\end{tabular}
\caption{\textbf{Ablation study on main components:} distillation (D), unprojected points filtering (M), and noise (N).}\label{tab:ablation}\vspace{-5pt}
\end{table}
\begin{table}[t]
\centering
\begin{tabular}{l|cccc|c} 
\hlinewd{0.8pt}
{Methods}           & Abs & Sq & RMSE & RMSElog & $\delta$ < 1.25  \\ 
\hline
Ours (sup.)         &\textbf{0.101}     &\textbf{0.652}     &4.264      &\textbf{0.176}     &0.891      \\
Ours (unsup.)       &\textbf{0.101}     &0.657     &\textbf{4.262}      &\textbf{0.176}       &\textbf{0.892}      \\
\hlinewd{0.8pt}
\end{tabular}
\caption{\textbf{Evaluation of final depth quality.}}\label{tab:final}\vspace{-10pt}
\end{table}

\textbf{Results on Cityscape.} In \tabref{tab:cittyscape} and \figref{city}, we also evaluated our
method on the Cityscapes dataset~\cite{cordts2016cityscapes}, where we evaluate the generalization ability of the networks learned from the KITTI dataset~~\cite{geiger2012we}. The evaluation is performed with the depth by the Semi-Global Matching (SGM)~\cite{hirschmuller2005accurate} as a ground-truth. The outstanding performance of our method demonstrates the proposed mutual distillation shows a satisfactory generalization capability for different datasets. 

\subsection{Ablation Study}
\textbf{Comparison of Mutual Distillation Modules.}
We analyze four different kinds of techniques to define our mutual distillation loss. We consider the uncertainty learned in our method, and the confidence learned from CCNN~\cite{poggi2016learning}, as done in~\cite{choi2020adaptive}. We first evaluate uncertainty weighting (UW) in our method, and uncertainty thresholding (UT) (with parameter $\tau = 0.1$), and both (UWT). In addition, we evaluate the confidence thresholding (CT), used in stereo knowledge distillation framework~\cite{ godard2017unsupervised, guo2018learning,cho2019large,tonioni2019unsupervised}. In this study, our uncertainty weighting (UW) for mutual distillation shows the best performance. We hypothesize that since our uncertainty was simultaneously trained with depth prediction, its magnitude itself can be reliable cue to adjust the loss when transferring depth knowledge, while confidence learned from additional networks, e.g., CCNN~\cite{poggi2016learning} needs the proper thresholding, which is hard to tune.

\textbf{Components Analysis.} To better understand how the components of our model contribute to the overall performance, in \tabref{tab:ablation}, we conduct ablation study on key components, mutual distillation (D), unprojected proint filtering (M), and noise (N). The results demonstrate that the baseline model, without any our contributions, performs the worst. However, when combined together, all of our components lead to a significant improvement.

\textbf{Final Depth.} \tabref{tab:final} demonstrates the final performance, after convergence of training, of our two branches, supervised branch and unsupervised branch. Because each branch has its own loss functions, the performance may be different during training, but after convergence, the performance gap was marginal thanks to the proposed mutual distillation loss functions. Empirically, we select the depths from unsupervised branch as our final depth prediction results as they are more plausible.

\section{Conclusion}
In this paper, we have proposed a novel semi-supervised learning (SSL) framework for monocular depth estimation. To achieve the complementary advantages of both sparse supervised and unsupervised loss functions for monocular depth estimation, we defined two separate branches for each loss, and distilled the depth knowledge each other with the learned uncertainties. We also proposed to apply different augmentation to each network, which boosts the robustness of the networks. We have shown that our method surpasses the current state-of-the-art in several benchmarks.

\noindent\textbf{Acknowledgements.}
This research was supported by the MSIT, Korea (IITP-2022-2020-0-01819, ICT Creative Consilience program), and National Research Foundation of Korea (NRF-2021R1C1C1006897). 

\newpage
\bibliographystyle{IEEEtran}
\bibliography{reference}

\end{document}